\documentclass[acmtog,authorversion]{acmart}
\acmSubmissionID{250}

\usepackage{arydshln}
\usepackage{soul}
\usepackage{url}
\usepackage{booktabs}
\usepackage{multirow}
\usepackage{float}
\usepackage{dblfloatfix}
\usepackage{wrapfig}
\usepackage[subtle]{savetrees}

\setcitestyle{square}

\usepackage[ruled]{algorithm2e} 

\SetAlFnt{\small}
\SetAlCapFnt{\small}
\SetAlCapNameFnt{\small}
\SetAlCapHSkip{0pt}
\IncMargin{-\parindent}

\newcommand\T{\rule{0pt}{2.6ex}}       
\newcommand\B{\rule[-1.2ex]{0pt}{0pt}} 

\newcommand{\parahead}[1]{\textbf{#1}:\ }

\newcommand{\change}[1]{{{#1}}}
\newcommand{\ignore}[1]{{}}

\usepackage{capt-of}
\newcommand{\etal}{et al.}
\newcommand{\eg}{e.g.~}
\newcommand{\ie}{i.e.,~}

\newcommand\blfootnote[1]{%
  \begingroup
  \renewcommand\thefootnote{}\footnote{#1}%
  \addtocounter{footnote}{-1}%
  \endgroup
}

\setcopyright{acmlicensed}\acmJournal{TOG}
\acmYear{2020}\acmVolume{39}\acmNumber{4}\acmArticle{1}\acmMonth{7} \acmDOI{10.1145/3386569.3392410}

\begin{document}
\title{XNect: Real-time Multi-Person 3D Motion Capture with a Single RGB Camera}

\author{
Dushyant Mehta$^{1,2}$, Oleksandr Sotnychenko$^{1,2}$, Franziska Mueller$^{1,2}$, Weipeng Xu$^{1,2}$, Mohamed Elgharib$^{1,2}$, Pascal Fua$^{3}$, Hans-Peter Seidel$^{1,2}$, Helge Rhodin$^{3,4}$, Gerard Pons-Moll$^{1,2}$, Christian Theobalt$^{1,2}$
}

\affiliation{%
\\
  \institution{
    $^{1}$Max Planck Institute for Informatics, $^{2}$Saarland Informatics Campus,
     $^{3}$EPFL, 
     $^{4}$University of British Columbia
  }
}
\renewcommand\shortauthors{Mehta, D. et al.}
\begin{teaserfigure}
  \includegraphics[width=\linewidth]{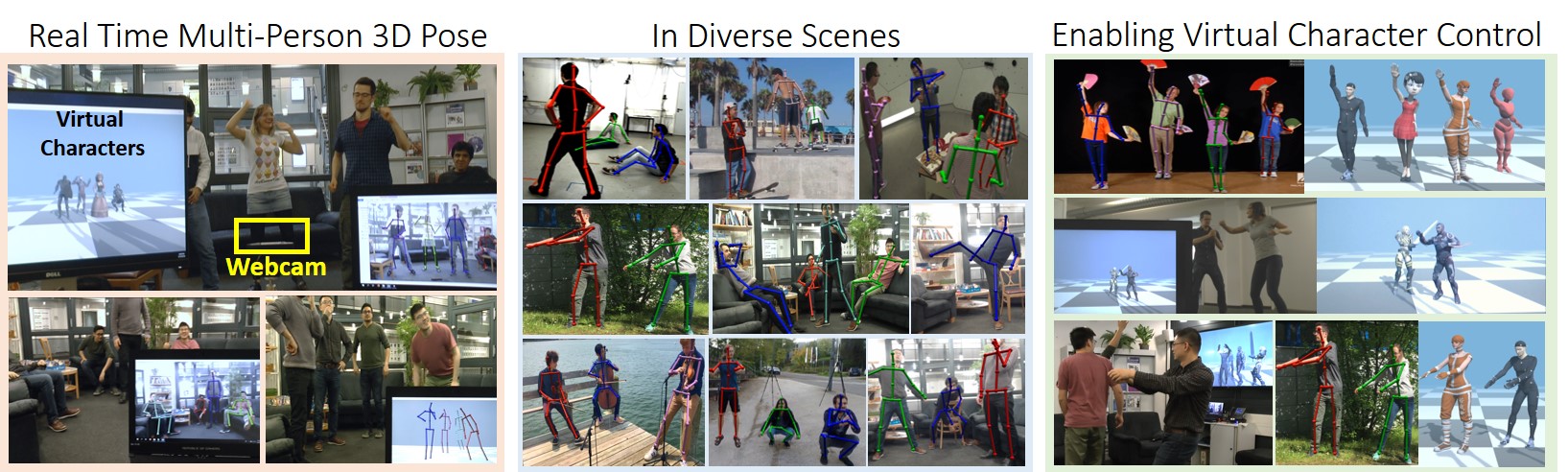}
  \caption
  {Our real-time monocular RGB based 3D motion capture provides temporally coherent estimates of the full 3D pose of multiple people in the scene, handling occlusions and interactions in general scene settings, and localizing subjects relative to the camera. Our design allows the system to handle large groups of people in the scene with the run-time only minimally affected by the number of people in the scene. Our method yields full skeletal pose in terms of joint angles, which can readily be employed for live character animation. Some images courtesy KNG Music (\url{https://youtu.be/_xCKmEhKQl4}), Music Express Magazine (\url{https://youtu.be/kX6xMYlEwLA}). 3D characters from Mixamo~\cite{mixamo}.}
  \label{fig:teaser}
\end{teaserfigure}

\begin{abstract}
We present 
a real-time approach for multi-person 3D motion capture at over $30$~fps using a single RGB camera. 
\ignore{It operates in generic scenes and is robust to difficult occlusions both by other people and objects.  }
It operates successfully in generic scenes which may contain occlusions by objects and by other people.  
Our method operates in subsequent stages.
The first stage is a convolutional neural network (CNN) that estimates 2D and 3D pose features along with identity assignments for all visible joints of all individuals.
We contribute a new architecture for this CNN, called \textit{SelecSLS Net}, that uses novel selective long and short range skip connections to improve the information flow allowing for a drastically faster network without compromising accuracy.
In the second stage, a fully-connected neural network turns the possibly partial (on account of occlusion) 2D pose and 3D pose features for each subject into a complete 3D pose estimate per individual.
The third stage applies space-time skeletal model fitting to the predicted 2D and 3D pose per subject to further reconcile the 2D and 3D pose, and enforce temporal coherence. %
Our method returns the full skeletal pose in joint angles for each subject. 
This is a further key distinction from previous work that do not produce joint angle results of a coherent skeleton in real time for multi-person scenes. 
The proposed system runs on consumer hardware at a previously unseen speed of more than $30$~fps given 512x320 images as input while achieving state-of-the-art accuracy, which we will demonstrate on a range of challenging real-world scenes. 

\end{abstract}

%
%



\thanks{This work was funded by the ERC Consolidator Grant 4DRepLy (770784). We thank Gereon Fox, Abhimitra Meka, Ikhsanul Habibie, Vikram Singh, Marc Habermann, Neng Qian, Jiayi Wang, Vladislav Golyanik, Edgar Tretschk, Tarun Yenamandra, Lingjie Liu, and Ayush Tewari for helping with data capture and the preparation of video materials.}

\maketitle
\section{Introduction}
\label{sec:intro}

\blfootnote{Project Website: \url{http://gvv.mpi-inf.mpg.de/projects/XNect/}}

Optical human motion capture is a key enabling technology in visual computing and related fields~\cite{chai2005performance,starck2007surface,Menache:2010:UMC:1965363}. 
For instance, it is widely used to animate virtual avatars and humans in VFX. It is a key component of many man-machine interfaces and is central to biomedical motion analysis. 
In recent years, computer graphics and computer vision researchers have developed new motion capture algorithms that operate on ever simpler hardware and under far less restrictive constraints than before.   
These algorithms do not require special body suits, dense camera arrays, in-studio recording, or markers. Instead, they only need a few calibrated cameras to capture people wearing everyday clothes outdoors, \eg\citet{stoll_fast_iccv2011,RhodinRCRST16,elhayek_convmocap_TPAMI2016,omran2018neural,kanazawa2018endtoend,pavlakos2019expressive,xiang2018monocular,VNect_SIGGRAPH2017,huang2017towards,fang2018learning}.
The latest approaches leverage the power of deep neural networks to capture 3D human pose from a single color image, opening the door to many exciting applications in virtual and augmented reality. Unfortunately, the problem remains extremely challenging due to depth ambiguities, occlusions, and the large variety of appearances and scenes. 

More importantly, most methods fail under occlusions and focus on a single person. \change{Some recent methods instead focus on the egocentric setting~\cite{rhodin_egocap_SIGGRAPHAsia2016,tome2019xr,xu2019mo}.}
Single person tracking \change{in the outside-in setting (non-egocentric)} is already hard and starkly under-constrained; multi-person tracking is incomparably harder due to mutliple occlusions, challenging body part to person assignment, and is computationally more demanding.
This presents a practical barrier for many applications such as gaming and social VR/AR, which require tracking \emph{multiple people} from low cost sensors, and in \emph{real time}.

Prior work on multi-person pose estimation runs at best at interactive frame rates (10-15 fps)~\citep{rogez_lcrpp,dabral2019multi} or offline~\citep{Moon_2019_ICCV_3DMPPE}, and produces per-frame joint position estimates which cannot be directly employed in many end applications requiring joint angle based avatar animations. 

We introduce a real-time algorithm for motion capture of multiple people in common interaction scenarios using a single color camera. Our full system produces the skeletal joint angles of multiple people in the scene, along with estimates of 3D localization of the subjects in the scene relative to the camera. Our method operates at more than 30 frames-per-second and delivers state-of-the-art accuracy and temporal stability. Our results are of a similar quality as commercial depth sensing based mocap systems.

To this end, we propose a new pose formulation and a novel neural network architecture, which jointly enable real-time performance, while handling inter-person and person-object occlusions. A subsequent model-based pose fitting stage produces temporally stable 3D skeletal motions. 
Our pose formulation uses two deep neural network stages that perform local (per body joint) and global (all body joints) reasoning, respectively. \textit{Stage I} is fully convolutional and jointly reasons about the 2D and 3D pose for all the subjects in the scene at once, which ensures that the computational cost does not increase with the number of individuals. 
Since \textit{Stage I} handles the already complex task of parsing the image for body parts, as well as associating the body parts to identities, our key insight with regards to the pose formulation is to have \textit{Stage I} \emph{only consider body joints for which direct image evidence is available}, i.e., joints that are themselves visible or their kinematic parents are visible. This way \textit{Stage I} does not have to spend representational capacity in hallucinating poses for joints that have no supporting image evidence. 
For each visible body joint, we predict the 2D part confidence maps, information for associating parts to an individual, and an \emph{intermediate 3D pose encoding} for the bones that connect at the joint. Thus, the 3D pose encoding is only cognizant of the joint's immediate neighbours (\textit{local}) in the kinematic chain.  
A compact fully-connected network forms \textit{Stage II}, which relies on the intermediate pose encoding and other evidence extracted in the preceding stage, to decode the complete 3D pose. 
The \textit{Stage II} network 
is able to reason about occluded joints using the full body context (\emph{global}) for each detected subject, and leverages learned pose priors, and the subject 2D and 3D pose evidence.
This stage is compact, highly efficient, and acts in parallel for all detected subjects. 

\textit{Stage I} is the most computationally expensive part of our pipeline, and the main bottleneck in achieving real-time performance.

We achieve real-time performance by contributing a new convolutional neural network (CNN) architecture in \textit{Stage I} to speed up the most computationally expensive part of our pipeline. We refer to the new architecture as \textit{SelecSLS Net}. Our proposed architecture depends on far fewer features than competing ones, such as ResNet-50~\cite{he_resnet_cvpr2016}, without any accuracy loss thanks to our insights on selective use of short and long range concatenation-skip connections. This enables fast inference on the complete input frame, without the added pre- or post-processing complexity of a separate bounding box tracker for each subject.
Further, the compactness of our \textit{Stage II} network, which reconciles the partially incomplete 2D pose and 3D pose encoding to a full body pose estimate, enables it to simultaneously handle many people with minimal overhead on top of \textit{Stage I}.
Additionally, we fit a model-based skeleton to the 3D and 2D predictions in order to satisfy kinematic constraints and reconcile the 2D and 3D predictions across time. This produces temporally stable predictions, with skeletal angle estimates, which can readily 
drive virtual characters.

In summary, our technical innovations and new design insights at the individual stages, as well as our insights guiding the proposed multi-stage design enable our final contribution: a complete algorithm for multi-people 3D motion capture from a single camera that achieves real-time performance without sacrificing reliability or accuracy. The run time of our system only mildly depends on the number of subjects in the scene, and even crowded scenes can be tracked at high frame rates. We demonstrate our system's performance on a variety of challenging multi-person scenes.

\section{Related Work}
\label{sec:rel-work}
We focus our discussion on relevant 2D and 3D human pose estimation from monocular RGB methods, in both single- and multi-person scenarios--for overview articles refer to \citet{sarafianos_posesurvey_cviu2016,DBLP:journals/jcst/XiaGLYC17}.
We also discuss prior datasets, and neural network architectures that inspired ours.

\parahead{Multi-Person 2D Pose Estimation}
Multi-person 2D pose estimation methods can be divided into bottom-up and top-down approaches. Top-down approaches first detect individuals in a scene and fall back to single-person 2D pose approaches or variants for pose estimation~\cite{pishchulin_reshape_cvpr12,gkioxari2014using,sun2011articulated,iqbal2016multi,papandreou2017towards}. Reliable detection of individuals under significant occlusion, and tracking of people through occlusions remains challenging.

Bottom-up approaches instead first localize the body parts of all subjects and associate them to individuals in a second step. Associations can be obtained by predicting joint locations and their identity embeddings together~\cite{newell_associative_nips17}, or by solving a graph cut problem~\cite{pishchulin_deepcut_cvpr16,insafutdin_arttrack_cvpr17}. This involves solving an NP-hard integer linear program which easily takes hours per image. The work of~\citet{insafutdin_arttrack_cvpr17} improves over~\citet{pishchulin_deepcut_cvpr16} by including image-based pairwise terms and stronger detectors based on ResNet~\cite{he_resnet_cvpr2016}. This way reconstruction time reduces to several minutes per frame. 
\citet{cao_affinity_2017} predict joint locations and part affinities (PAFs), which are 2D vectors linking each joint to its parent. PAFs allow quick and greedy part association, enabling real time mutli-person 2D pose estimation. 
Our \textit{Stage I} uses similar ideas to localize and assign joints in 2D, but we also predict an intermediate 3D pose encoding per joint which enables our subsequent stage to produce accurate 3D body pose estimates.
\citet{GulerNK18} compute dense correspondences from pixels to the surface of SMPL~\shortcite{smpl2015loper}, but they do not estimate 3D pose.%

\parahead{Single-Person 3D Pose Estimation}
 Monocular single person 3D pose estimation was previously approached with generative methods using physics priors~\cite{Wei:2010:VideoMocap}, or semi-automatic analysis-by-synthesis fitting of parametric body models~\cite{Jain:2010:MovieReshape,Guan2009}.
Recently, methods employing CNN based learning approaches led to important progress ~\cite{ionescu_human36_pami14,sigal_humaneva_ijcv10,tekin_structured_bmvc16,pavlakos_volumetric_cvpr17,li_maximum_iccv2015,li_accv14,sun2017compositional,sun2018integral}. 
These methods can broadly be classified into direct regression and `lifting' based approaches. Regressing straight from the image requires large amounts of 3D-pose labelled images, which are difficult to obtain.
Therefore, existing datasets are captured in studio scenarios with limited pose and appearance diversity~\cite{ionescu_human36_pami14}, or combine real and synthetic imagery~\cite{Deep3DPose}.
 Consequently, to address the 3D data scarcity, transfer learning using features learned on 2D pose datasets has been applied to improve 3D pose estimation~\cite{VNect_SIGGRAPH2017,mehta_mono_3dv17,popa2017deep,zhou2017towards,sun2017compositional,tekin_fusion_arxiv16}.
 
`Lifting' based approaches predict the 3D pose from a separately detected 2D pose~\cite{martinez20173dbaseline}. This has the advantages that 2D pose datasets are easier to obtain in natural environments, and the lifting can be learned from MoCap data without overfitting on the studio conditions. While this establishes a surprisingly strong baseline, lifting is ill-posed and body-part depth disambiguation is often not possible without additional cues from the image. Other work has proposed to augment the 2D pose with relative depth ordering of body joints as additional context to disambiguate 2D to 3D lifting~\cite{posebits_cvpr14,pavlakos2018ordinal}.
Our approach can be seen as a hybrid of regression and lifting methods: An encoding of the 3D pose of the visible joints is regressed directly from the image (Stage I), with each joint only reasoning about its immediate kinematic neighbours (local context). This encoding, along with 2D joint detection confidences augments the 2D pose and is `decoded' into a complete 3D body pose by \textit{Stage II} reasoning about all body joints (global context).

Some recent methods integrate a 3D body model~\cite{smpl2015loper} within a network, and train using a mixture of 2D poses and 3D poses to predict 3D pose and shape from single images~\cite{omran2018neural,pavlakos2018humanshape,kanazawa2018endtoend,tung2017self}. Other approaches optimize a body model or a template~\cite{xu2018monoperfcap,habermann2019TOG} to fit 2D poses or/and silhouettes~\cite{bhatnagar2019mgn,thiemo2018,thiemo2018_3DV,alldieck19cvpr,bogo_smpl_eccv16,Lassner:UP:2017,kolotouros2019convolutional,kolotouros2019learning,guler2019holopose}. Very few are able to work in real time, and none of them handles multiple people.

Prior real-time 3D pose estimation approaches \cite{VNect_SIGGRAPH2017} designed for single-person scenarios fail in multi-person scenarios. Recent offline single-person approaches~\cite{humanMotionKanazawa19} produce temporally coherent sequences of SMPL~\shortcite{smpl2015loper} parameters, but work only for unoccluded single-person scenarios. In contrast, our proposed approach runs in real time for multi-person scenarios, and produces temporally coherent joint angle estimates at par with offline approaches, while successfully handling object and inter-person occlusions. 

\begin{figure*}[t]
  \includegraphics[width=\linewidth]{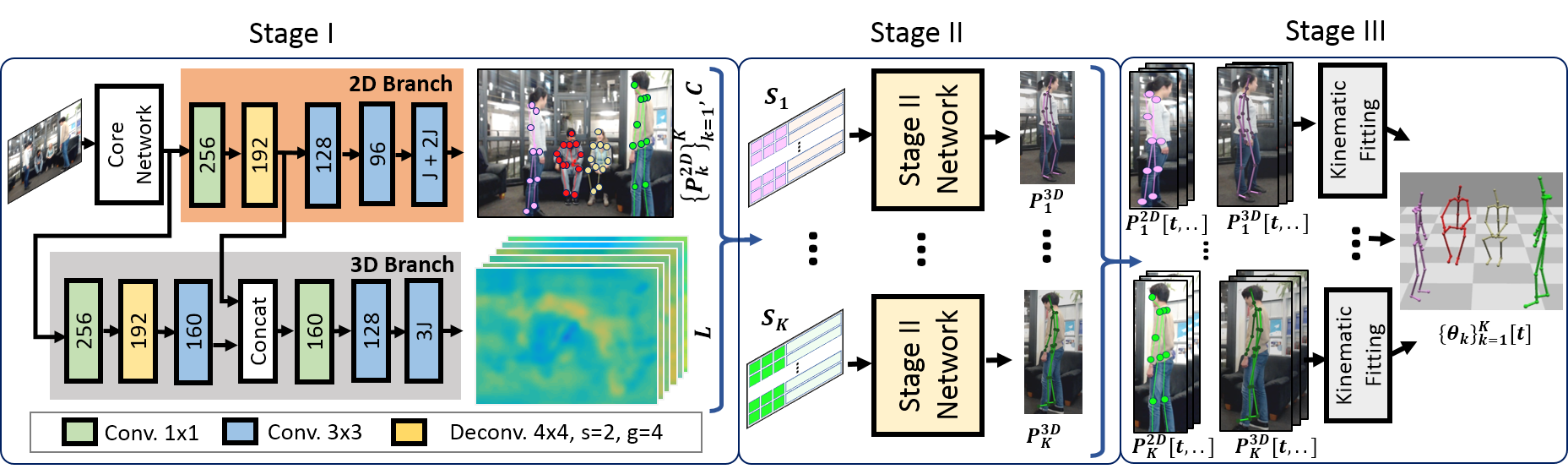}
  \caption
  {\textbf{Overview:} Computation is separated into three stages, the first two respectively performing per-frame local (per body joint) and global (all body joints) reasoning, and the third performing temporal reasoning across frames: 
  \textit{Stage I} infers 2D pose and intermediate 3D pose encoding for visible body joints, using a new \textit{SelecSLS Net} architecture. The 3D pose encoding for each joint only considers local context in the kinematic chain. \textit{Stage II} is a compact fully-connected network that runs in parallel for each detected person, and reconstructs the complete 3D pose, including occluded joints, by leveraging global (full body) context. \textit{Stage III} provides temporal stability, localization relative to the camera, and a joint angle parameterization through kinematic skeleton fitting.
}
  \label{fig:pipeline}
\end{figure*}

\parahead{Multi-Person 3D Pose}
Earlier work on monocular multi-person 3D pose capture often followed a generative formulation, \eg estimating 3D body and camera pose from 2D landmarks using a learned pose space~\cite{ramakrishna2012reconstructing}.
We draw inspiration from and improve over limitations of recent deep learning-based methods.
\citet{rogez_lcr_cvpr17} 
use a Faster-RCNN~\shortcite{ren_faster_rcnn_nips15} based approach and first find representative poses of discrete pose clusters that are subsequently refined.
The LCRNet++ implementation of this algorithm uses a ResNet-50 base network 
and achieves non-real-time interactive $10-12$fps on consumer hardware even with the faster but less accurate `demo' version that uses fewer anchor poses.
\citet{dabral2019multi} use a similar Faster-RCNN based approach, and predict 2D keypoints for each subject. Subsequently, the predicted 2D keypoints are lifted to 3D pose. We show that incorporating additional information, such as the keypoint confidence, and 3D pose encodings in the `lifting' step results in a much higher prediction accuracy. 
\citet{Moon_2019_ICCV_3DMPPE} employ a prior person detection step, and pass resized image crops of each detected subject to the pose estimation network. As prior work~\citep{cao_affinity_2017} has shown, such an approach results high pose estimation accuracy, but comes at the cost of a significant increase in inference time. Not only does such an approach work at offline rates, the per-frame inference time scales with the number of subjects in the scene, making it unsuitable for real-time applications.

The aforementioned detection based approaches predict multiple proposals per individual and fuse them afterwards. This is time consuming, and in many cases it can either incorrectly merge nearby individuals with similar poses, or fail to merge multiple proposals for the same individual. Beyond the cost and potential errors from fusing pose estimates, multiple detections of the same subject further increase the inference time for the approach of \citet{Moon_2019_ICCV_3DMPPE}. 

Our approach, being bottom-up, does not produce multiple detections per subject. The bottom-up approach of 
\citet{mehta_3dv18} 
predicts the 2D and 3D pose of all individuals in the scene using a fixed number of feature maps, which jointly encode for any number of individuals in the scene. 
This introduces potential \emph{conflicts} when subjects overlap, for which a complex encoding and read-out scheme is introduced.
The 3D encoding treats each limb and the torso as distinct objects, and encodes the 3D pose of each `object' in the feature maps at the pixel locations corresponding to the 2D joints of the `object'. The encoding can thus handle partial inter-personal occlusion by dissimilar body parts. Unfortunately, the approach still fails when similar body parts of different subjects overlap.
Similarly, \citet{zanfir_nips18} jointly encode the 2D and 3D pose of all subjects in the scene using a fixed number of feature maps. 
Different from ~\citet{mehta_3dv18}, they encode the full 3D pose vector at all the projected pixels of the skeleton, and not just at the body joint locations, which makes the 3D feature space rife with potential encoding conflicts. For association, they learn a function to evaluate limb grouping proposals. 
A 3D pose decoding stage extracts 3D pose features per limb and uses an attention mechanism to combine these into a 3D pose prediction for the limb. 

One of our key contributions 
is to only reason about body joints for which there is direct image evidence available, i.e., the joint itself, or its parent/child is visible. A subsequent compact fully-connected network can decode this potentially incomplete information to the full 3D pose. 
Such a hybrid of image-to-pose regression and 2D-to-3D lifting helps overcome the limitations of the individual approaches. 
The 3D pose encodings are a strong cue for the 3D pose in the absence of conflicts, whereas the global context in \textit{Stage II} and the 2D pose help resolve the very-limited conflicts when they occur, and use learned pose priors to fill in the missing body joints. 
In contrast, the encodings of~\citet{zanfir_nips18} and~\citet{mehta_3dv18} encode the pose for all joints in the full body or limb-wise, respectively, regardless of available image evidence for each joint, making the already difficult task more difficult, and having several avenues of potential encoding conflicts.
Furthermore, we impose kinematic constraints with a model based fitting stage, which also allows for temporal smoothness.
The approach of \citet{zanfir2018monocular} also combines learning and optimization, but their space-time optimization over all frames is not real-time.

Different from prior approaches, our approach works in real-time at $25-30$~fps using a single consumer GPU, yielding skeletal joint angles and camera relative positioning of the subject, which can be readily be used to control animated characters in a virtual environment.
Our approach predicts the complete body pose even under significant person-object occlusions, and is more robust to inter-personal occlusions. 

\parahead{3D Pose Datasets}
There exist many datasets with 3D pose annotations in single-person scenarios~\cite{ionescu_human36_pami14,sigal_humaneva_ijcv10,trumble2017total,vonPon2016a,mehta_mono_3dv17} or multi-person with only 2D pose annotations~\cite{andriluka_mpii2d_cvpr14,lin_coco_eccv14}. As multi-person 3D pose estimation started to receive more attention, datasets such as MarCOnI~\cite{elhayek_convmocap_TPAMI2016} with a lower number of scenes and subjects, and the more diverse Panoptic~\cite{jooICCV2015} and MuCo-3DHP~\cite{mehta_3dv18} datasets have come about. 
LCRNet~\cite{rogez_lcr_cvpr17} uses 2D to 3D lifting to create pseudo annotations on the MPII 2D pose dataset~\cite{andriluka_mpii2d_cvpr14}, and LCRNet++~\cite{rogez_lcrpp} uses synthetic renderings of humans from a multitude of single person datasets.

Recently, the 3D Poses in the Wild (3DPW) dataset~\cite{vonMarcard2018} features multiple people outdoors recorded with a moving camera and includes ground truth 3D pose. The number of subjects with ground truth pose is however limited. To obtain more variation in training, we use the recently published MuCo-3DHP~\cite{mehta_3dv18}, which is a multi-person training set of composited real images with 3D pose annotations from the single person MPI-INF-3DHP~\shortcite{mehta_mono_3dv17} dataset.

\parahead{Convolutional Network Designs} 
ResNet~\cite{he_resnet_cvpr2016} and derivatives~\cite{xie2017aggregated} incorporate explicit information flowing from earlier to later feature layers in the network through summation-skip connections. 
This permits training of deeper and more powerful networks.
Many architectures based on this concept have been proposed, such as Inception~\cite{szegedy2017inception} and ResNext~\cite{xie2017aggregated}. 

Because increased depth and performance comes at the price of higher computation times during inference, \change{as well as a higher number of parameters}, specialized architectures for faster test time computation were proposed, such as AmoebaNet~\cite{real2018regularized}, MobileNet~\cite{sandler2018mobilenetv2,howard2019searching}, ESPNet~\cite{mehta2018espnet}, ERFNet \cite{romera2018erfnet}, \change{EfficientNet~\cite{tan2019efficientnet}, as well as architectures such as SqueezeNet\cite{SqueezeNet} that target parameter efficiency.} These are however not suited for our use case for various reasons: 
Many architectures with depthwise convolutions are optimized for inference on specific edge devices~\cite{sandler2018mobilenetv2}, and lose accuracy in lieu of speed. Increasing the width or depth of these networks \cite{howard2019searching,tan2019efficientnet} to bring the accuracy closer to that of vanilla ResNets results in GPU runtimes comparable to typical ResNet architectures.
ESPNet uses hierarchical feature fusion but 
produces non-smooth output maps with grid artifacts due to the use of dilated convolutions. These artifacts impair part association performance in our pose estimation setting.
\change{ShuffleNet~\cite{zhang2018shufflenet,ma2018shufflenet} use group convolutions and depthwise convolutions, and shuffle the channels between layers to promote information flow between channel groups.}
DenseNet~\cite{huang2017densely} 
uses full dense concatenation-skip connectivity, which results in a parameter efficient network but is slow due to the associated memory cost of the enormous number of concatenation operations.
\change{Recent work has also proposed highly computation-efficient networks~\cite{WangSCJDZLMTWLX19,SunXLW19} which maintain high to low resolution feature representations throughout the network, and do not lose accuracy in lieu of computational efficiency. However, the theoretical computational efficiency does not translate to computational speed in practice, and the models are up to twice as slow as ResNet networks for a given accuracy level. Approaches targeting parameter efficiency often do not result in computational speedups because either the computational cost is still high~\cite{SqueezeNet}, or the non-structured sparse operations resulting from weight pruning~\cite{frankle2018lottery} cannot be executed efficiently on current hardware.}
%

The key novel insight behind our proposed CNN architecture is the use of selective long-range and short-range concatenation-skip connections rather than the dense connectivity pattern of DenseNet. 
This results in a network significantly faster than ResNet-50 while retaining the same level of accuracy, avoids the artifacts and accuracy deficit of ESPNet, and eliminates the memory (and hence speed) bottlenecks associated with DenseNet.

\section{Method Overview}
\label{sec:method}
The input to our method is a live video feed, \ie a stream of monocular color frames showing a multi-person scene.
Our method has three subsequent stages, as shown in Fig.~\ref{fig:pipeline}. 
In Section~\ref{sec:per_frame_pose_est}, we discuss the first two stages, which together produce 2D and 3D pose estimates per frame.

\textit{Stage I} uses a convolutional neural network to process the complete input frame, jointly handling all subjects in the scene.
The \textit{Stage I} CNN predicts 2D body joint heatmaps, Part Affinity Fields to associate joints to individuals in the scene, and an intermediate 3D pose encoding per detected joint. 
After grouping the 2D joint detections from the first stage into individuals following the approach of~\cite{cao_affinity_2017}, 3D pose encodings per individual are extracted at the pixel locations of the visible joints and are input to the second stage together with the 2D locations and detection confidences of the individual's joints. 
\textit{Stage I} only reasons about visible body joints, and the 3D pose encoding per joint only captures the joint's pose relative to its immediate kinematic neighbours. The 3D pose encoding is discussed in Section \ref{sec:pose_formulation}. 

\textit{Stage II}, which we discuss in Section~\ref{sec:second_stage}, uses a lightweight fully-connected neural network that `decodes' the input from the previous stage into a full 3D pose, i.e. root-relative 3D joint positions for visible and occluded joints, per individual.  
This network incorporates 2D pose and 3D pose encoding evidence over all visible joints and an implicitly learned prior on 3D pose structure, which allows it to reason about occluded joints and correct any 3D pose encoding conflicts. 
A further advantage of a separate stage for full 3D pose reasoning is that it allows the use of a body joint set different from that used for training \textit{Stage I}.
In our system, 3D pose inference of \textit{Stage I} and \textit{Stage II} can be parallelized on a GPU, with negligible dependence of inference time on the number of subjects. 

\textit{Stage III}, discussed in Section \ref{sec:skeleton_fitting}, performs sequential model fitting on the live stream of 2D and 3D predictions from the previous stages. 
A kinematic skeleton is fit to the history of per-frame 2D and root-relative 3D pose predictions to obtain temporally coherent motion capture results. 
We also track person identity, full skeletal joint angles, and the camera relative localization of each subject in real time.

\begin{figure}[t]
  \includegraphics[width=0.8\linewidth]{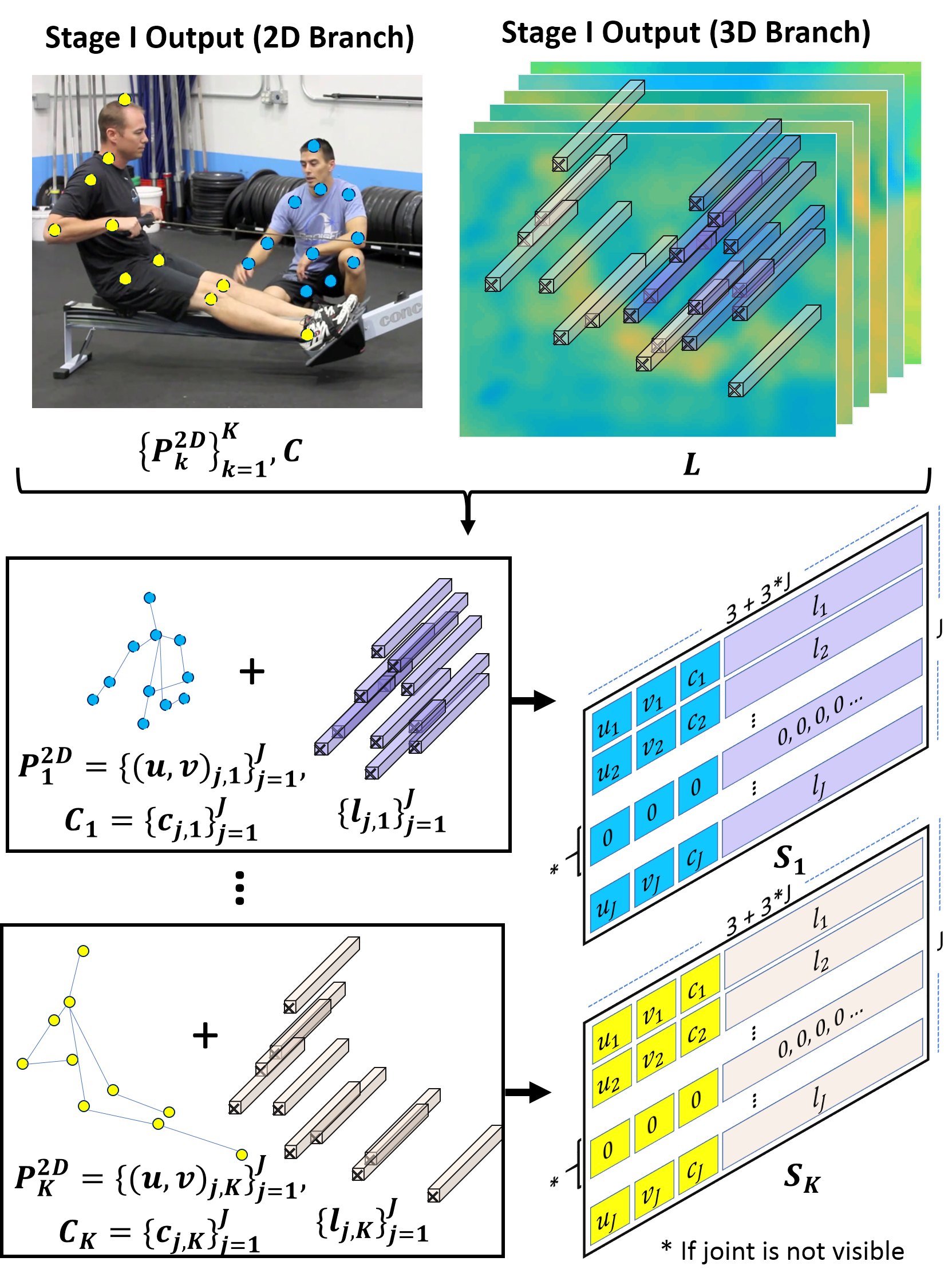}
  \caption
  {\textbf{Input to \textit{Stage II}:} $S_k$ for each detected individual $k$, is comprised of the individual's 2D joint locations $P_k^{2D}$, the associated joint detection confidence values $C$ extracted from the 2D branch output, and the respective 3D pose encodings $\{l_{j,k}\}_{j=1}^{J}$ extracted from the output of the 3D branch.
  Refer to Section~\ref{sec:per_frame_pose_est} for details.
  }
  \label{fig:feature_stage2}
  \vspace{-0.3cm}
\end{figure}
Our algorithm and pose representation applies to any CNN architecture suitable for keypoint prediction. However, to enable
%
fast inference on typical consumer-grade GPUs, we propose the novel \textit{SelecSLS Net} architecture for the backbone of \textit{Stage I} CNN.
It employs selective long and short range concatenation-skip connections to promote information flow across network layers which allows to use fewer features and have a much smaller memory footprint leading to a much faster inference time but comparable accuracy in comparison to ResNet-50. We discuss our contributions in that regard separately in Section~\ref{sec:dlnas}.

In Section~\ref{sec:results}, we present ablation and comparison studies, both quantitative and qualitative, and show applications to animated character control.

\section{Per-Frame Pose Estimation: \textit{Stage I} \& \textit{Stage II}}
\label{sec:per_frame_pose_est}

Given an image $I$ of dimensions $w \times h$ pixels, we seek to estimate the 3D pose $\{P_k^{3D}\}_{k=1}^{K}$ of the unknown number of $K$ individuals in the scene. 
$P_k^{3D} \in \mathbb{R}^{3\times J}$ represents the root (pelvis)-relative 3D coordinates of the $J$ body joints. 
The task is implemented in the first two stages of our algorithm, which we detail in the following.

\subsection{\textit{Stage I} Prediction}
\label{sec:fisrt_stage}

Our first stage uses a CNN that features an initial core (or backbone) network that splits into two separate branches for 2D pose prediction and 3D pose encoding, as shown in Figure~\ref{fig:pipeline}. The core network outputs features at $\frac{w}{16} \times \frac{h}{16}$ pixel spatial resolution, and uses our new proposed network design that offers a high accuracy at high runtime efficiency, which we detail in Section~\ref{sec:dlnas}.
The outputs of each of the 2D and 3D branches are at $\frac{w}{8} \times \frac{h}{8}$ pixels spatial resolution. 
The 3D pose branch also makes use of features from the 2D pose branch. We explain both branches and the \textit{Stage I} network training in the following.  

\subsubsection{2D Branch: 2D Pose Prediction and Part Association}
2D pose is predicted as 2D heatmaps $H = \{H_j \in \mathbb{R}^{\frac{w}{8} \times \frac{h}{8}}\}_{j=1}^{J} $, where each map represents the per-pixel confidence of the presence of body joint type $j$ jointly for all subjects in the scene. 
Similar to~\cite{cao_affinity_2017}, we use Part Affinity fields $F = \{F_j \in \mathbb{R}^{\frac{w}{8} \times \frac{h}{8} \times 2}\}_{j=1}^{J} $ to encode body joint ownership using a unit vector field that points from a joint to its kinematic parent, and spans the width of the respective limb. 
For an input image, these Part Affinity Fields can be used to detect the individuals present in the scene and the visible body joints, and to associate visible joints to individuals. 
If the neck joint (which we hypothesize is visible in most situations) of an individual is not detected, we discard that individual entirely from the subsequent stages. 
For $K$ detected individuals, this stage outputs the 2D body joint locations in absolute image coordinates $P_k^{2D} \in \mathbb{Z}_{+}^{2\times J}$. 
Further, we get an estimate of the detection confidence $c_{j,k}$ of each body part $j$ and person $k$ from the heatmap maximum.

\subsubsection{3D Branch: Predicting Intermediate 3D Pose Encoding}
\label{sec:pose_formulation}

The 3D branch of the \textit{Stage I} network uses the features from the core network and the 2D branch to predict 3D pose encoding maps $L = \{L_j \in R^{\frac{w}{8} \times \frac{h}{8} \times 3}\}_{j=1}^{J} $. 
The encoding at the spatial location of each visible joint only encapsulates its 3D pose relative to the joints to which it directly connects in the kinematic chain. 

The general idea of such an encoding map is inspired by the approaches of ~Mehta \etal~\shortcite{VNect_SIGGRAPH2017}, Pavlakos \etal~\shortcite{pavlakos_volumetric_cvpr17} which represent the 3D pose information of joints in output maps at the spatial locations of the 2D detections of the respective joints.

\begin{figure}[]
  \includegraphics[width=\linewidth]{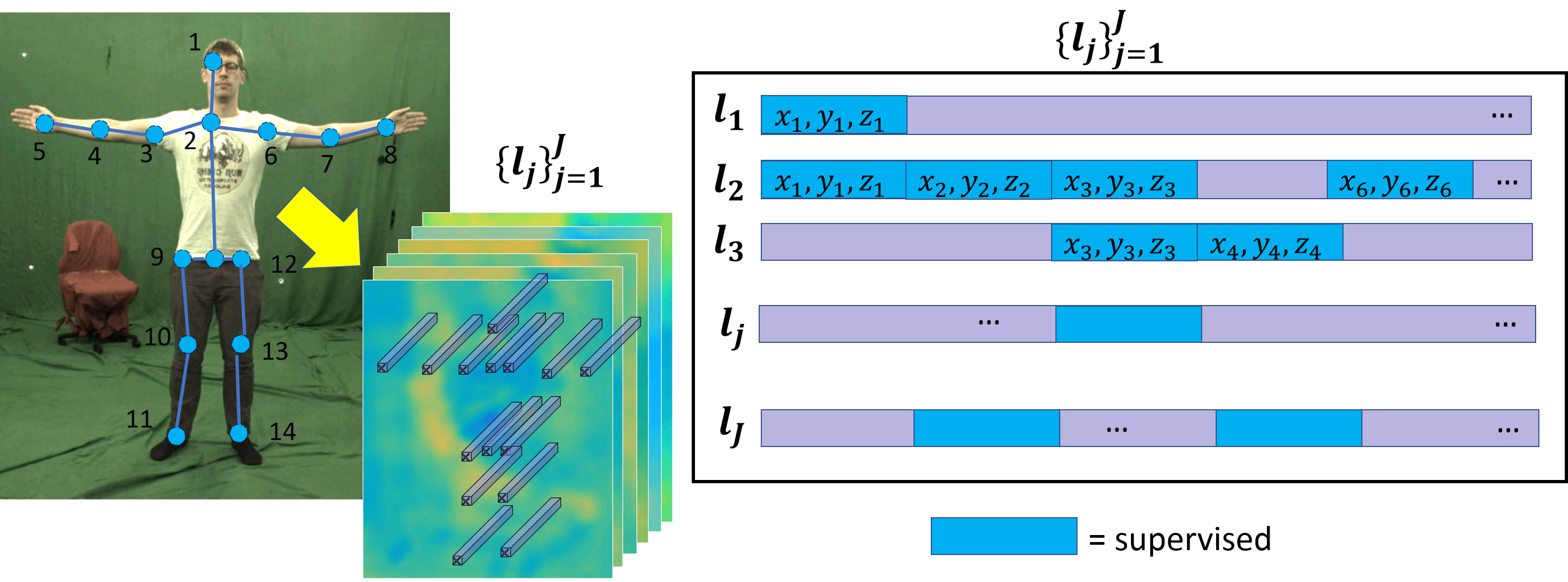}
  \caption
  {\textbf{3D Pose Encoding With Local Kinematic Context:} The supervision for the $1\times1\times(3 \cdot J)$ predicted 3D pose encoding vector $l_j$ \change{(shown on the right)} at each joint $j$ \change{(shown on the skeleton on the left)} is dependent on the type of the joint. $l_j$ only encodes the 3D pose information of joint $j$ relative to the joints it is directly connected to in the kinematic chain. This results in a channel-sparse supervision pattern as shown here, as opposed to each $l_j$ encoding the full body pose. The regions marked purple are not supervised and the network is free to predict any values there. \change{For example, `right shoulder' ($j=3$) connects to joints $j=2$ and $j=4$, so in $l_3$ the pose of $j=3$ relative to $j=2$ ($x_3,y_3,z_3$), and the pose of $j=4$ relative to $j=3$ ($x_4,y_4,z_4$) are supervised.} See Section~\ref{sec:pose_formulation}.}
  \label{fig:feat_supervision}
\end{figure}

Our specific encoding in $L$ works as follows: Consider the $1\times1\times(3 \cdot J)$ vector $l_{j,k}$ extracted at the pixel location $(u,v)_{j,k}$ from the 3D output maps $L$. Here $(u,v)_{j,k}$ is the location of body joint $j$ of individual $k$. 
This $1\times1\times(3 \cdot J)$ feature vector is of the dimensions of the full 3D body pose, where the kinematic parent-relative 3D locations of each joint reside in separate channels. Importantly however, and in contrast to previous work~\cite{mehta_3dv18,zanfir_nips18}, instead of encoding the full 3D body pose, or per-limb pose, at each 2D detection location $(u,v)_{j,k}$, we only encode the pose of the corresponding joint (relative to its parent) and the pose of its children (relative to itself). In other words, at each joint location $(u,v)_{j,k}$, we restrict the supervision of the encoding vector $l_{j,k}$ to the subset of channels corresponding to the bones that meet at joint $j$, parent-to-joint and joint-to-child in the kinematic chain. We will refer to this as channel-sparse supervision of $ \{l_{j,k}\}_{j=1}^{J}$, and emphasize the distinction from channel-dense supervision. 
Figure~\ref{fig:feat_supervision} shows examples for head, neck and right shoulder. 
Consequently, 3D pose information for all the visible joints of all subjects is still encoded in $L$, albeit in a spatially distributed manner, and each 2D joint location $(u,v)_{j,k}$ is used to extract its corresponding 3D bones of subject $k$.
Our motivation for such a pose encoding is that the task of parsing in-the-wild images to detect 2D body part heatmaps under occlusion and clutter, as well as grouping the body parts with their respective person identities under inter-personal interaction and overlap is already challenging. 
Reasoning about the full 3D pose, including occluded body parts, adds further complexity, which not only requires increased representation capacity (thus increasing the inference cost), but also more labelled training data, which is scarce for multi-person 3D pose. The design of our formulation responds to both of these challenges.  
Supervising only the 3D bones corresponding to each visible joint ensures that mostly local image evidence is used for prediction, where the full body context is already captured by the detected 2D pose. 
For instance, it should be possible to infer the kinematic-parent relative pose of the upper arm and the fore arm by looking at the region centered at the elbow. This means better generalization and less risk to overfit to dataset specific long-range correlations.

Further, our use of channel-sparse (joint-type-dependent) supervision of $l_{j,k}$ is motivated by the fact that convolutional feature maps cannot contain sharp transitions~\cite{mehta_3dv18}. In consequence, full poses of two immediately nearby people are hard to encode.
E.g., the wrist of one person being in close proximity in the image plane to the shoulder of another person would require the full pose of two different individuals to be encoded in possibly adjacent pixel locations in the output map. 
Such encoding conflicts often lead to failures of previous methods, as shown in Figure 5 in the supplemental document. 
In contrast, our encoding in $L$ does not lead to encoding conflicts when different joints of separate individuals are in spatial proximity or even overlap in the image plane, because supervision is restricted to the channels corresponding to the body joint type. 
Consequently, our target output maps are smoother without sharp transitions, and more suitable for representation by CNN outpus. 
In Section~\ref{sec:sparse_v_dense} we show the efficacy of channel-sparse supervision for $\{l_{j,k}\}_{j=1}^{J}$ over channel-dense supervision across various 2D and 3D pose benchmarks.
Importantly, unlike Zanfir \etal~\shortcite{zanfir_nips18} and Mehta \etal~\shortcite{mehta_3dv18}, the 2D pose information is not discarded, and is utilized as additional relevant information for 3D pose inference in \textit{Stage II}, allowing for a compact and fast network. 
This makes it more suited for a real-time system than, for instance, the attention-mechanism-based inference scheme of Zanfir \etal~\shortcite{zanfir_nips18}.

For each individual $k$, the 2D pose $P_k^{2D}$, joint confidences $\{c_{j,k}\}_{j=1}^{J}$, and 3D pose encodings $\{l_{j,k}\}_{j=1}^{J}$ at the visible joints are extracted and input to Stage II (Sec.~\ref{sec:second_stage}). Stage II uses a fully-connected decoding network that leverages the full body context that is available to it, to give the complete 3D pose with the occluding joints filled in. We provide details of Stage II in Section~\ref{sec:second_stage}.

\subsubsection{\textit{Stage I} Training}
The \textit{Stage I} network is trained in multiple stages. First the core network and the 2D pose branch are trained for single person 2D pose estimation on the MPII~\cite{andriluka_mpii2d_cvpr14} and LSP~\cite{johnson_lsp_bmvc10,johnson_lspet_cvpr11} single person 2D datasets. Then, using these weights as initialization, it is trained for multi-person 2D pose estimation on MS-COCO~\cite{lin_coco_eccv14}. Subsequently the 3D pose branch is added and the two branches are individually trained on crops from MS-COCO and MuCo-3DHP~\cite{mehta_3dv18}, with the core network seeing gradients from both datasets via the two branches. Additionally, the 2D pose branch sees supervision from MuCo-3DHP dataset via heatmaps of the common minimum joint set between MS-COCO and MuCo-3DHP. We found that the pretraining on multi-person 2D pose data before introducing the 3D branch is important.

\subsection{\textit{Stage II} Prediction}
\label{sec:second_stage}
\begin{figure}[t]
  \includegraphics[trim=0 1.0cm 0 0,width=0.75\linewidth]{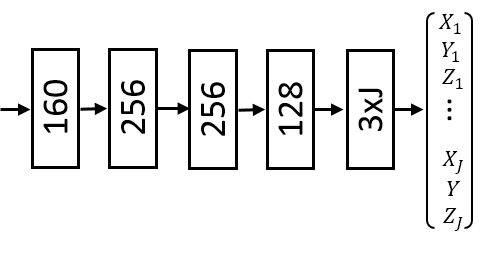}
  \caption
  {Lightweight fully connected network that forms \textit{Stage II} of our pipeline. The network decodes the inferred 2D body pose, joint detection confidences and 3D pose encodings coming from \textit{Stage I} to root-relative full body 3D pose $(X_j,Y_j,Z_j)$ estimates, leveraging full body context to fill in occluded joints.}
  \label{fig:stage2}
\end{figure}

\textit{Stage II} uses a lightweight fully-connected network to predict the root-relative 3D joint positions $\{P_k^{3D}\}_{k=1}^{K}$ for each individual considered visible after \textit{Stage I}. 
Before feeding the output from \textit{Stage I} as input, we convert the 2D joint position predictions $P_k^{2D}$ to a representation relative to the neck joint. 
%
For each individual $k$, at each detected joint location, we extract the $1\times1\times(3 \cdot J)$ 3D pose encoding vector $l_{j,k}$, as explained in the preceding section. 
The input to \textit{Stage II}, $S_k \in \mathbb{R}^{J\times(3+3 \cdot J)}$, is the concatenation of the neck relative $(u,v)_{j,k}$ coordinates of the joint, the joint detection confidence $c_{j,k}$ and the feature vector $l_{j,k}$, for each joint $j$. If the joint is not visible, we instead concatenate zero vectors of appropriate dimensions (see Figure~\ref{fig:feature_stage2}).
\textit{Stage II} comprises a 5-layer fully-connected network, which converts $S_k$ to a root-relative 3D pose estimate $P_k^{3D}$ (see Figure~\ref{fig:stage2}). 

We emphasize that unlike~\citet{mehta_3dv18}, we have a much sparser pose encoding $l_{j,k}$, and further only use it as a feature vector and not directly as the body part's pose estimate because jointly encoding body parts of all individuals in the same feature volume 
may result in corrupted predictions in the case of conflicts--same parts of different individuals in close proximity.
Providing the 2D joint positions and part confidences along with the feature vectors as input to the \textit{Stage II} network allows it to correct any conflicts that may arise. See Figure~5 in the supplemental document for a visual comparison of results against~\cite{mehta_3dv18}. 

 The inference time for \textit{Stage II} with a batch size of 10 is 1.6ms on an Nvidia K80, and 1.1ms on a TitanX (Pascal).

\subsubsection{\textit{Stage II} Training}
The \textit{Stage II} network is trained on uncropped frames from MuCo-3DHP~\cite{mehta_3dv18}. 
We run \textit{Stage I} on these frames and extract the 2D pose and 3D pose encodings. Then for each detected individual, we use the 
ground-truth root-relative 3D pose as the supervision target for $\{(X_j,Y_j,Z_j)\}_{j=1}^{J}$. 
Since the pose prediction can be drastically different from the ground truth when there are severe occlusions, we use the smooth-L1~\cite{ren_faster_rcnn_nips15} loss to mitigate the effect of such outliers.
In addition to providing an opportunity to reconcile the 3D pose predictions with the 2D pose, another advantage of a second stage trained separately from the first stage is that the output joint set can be made different from the joint set used for \textit{Stage I}, depending on which dataset was used for training \textit{Stage II} (joint sets typically differ across datasets).  
In our case, though there are no 2D predictions for foot tip, the 3D pose encoding for ankle encodes information about the foot tip, which is used in \textit{Stage II} to produce 3D predictions for foot tips.

\section{Sequential motion capture: \textit{Stage III}}
\label{sec:skeleton_fitting}
After \textit{Stage I} and \textit{Stage II} we have per-frame root-relative pose estimates for each individual.  
However, we have no estimates of person size or metric distance from the camera, person identities are not tracked across frames, and reconstructions are not in terms of joint angles.  
To remedy this, we infer and track person appearance over time, optionally infer absolute height from ground plane geometry, and fuse 2D and 3D predictions with temporal smoothness and joint limit constraints in a space-time kinematic pose fitting method. 

\subsection{Identity Tracking and Re-identification}
\label{sec:identity_tracking}
To distinguish poses estimated at distinct frames, we extend the previous pose notation with temporal indices in square brackets.
So far, per-frame 2D and 3D poses have been estimated for the current and past frames.
We need a fast method that maintains identity of a detected person across frames and re-identifies it after a period of full occlusion.   
To this end, we assign correspondences between person detections at the current timestep $t$, $\{P_{i}[t]\}_{i=1}^{K{[t]}}$, to the preceding ones $\{P_{k}[t-1]\}_{k=1}^{K{[t-1]}}$.
We model and keep track of person appearance with an HSV color histogram of the upper body region.
We discretize the hue and saturation channels into $30$ bins each and determine the appearance $A_{i[t]}$ as the class probabilities across the bounding box enclosing the torso joints in $\{P_{i}^{2D}[t]\}_i$.
This descriptor is efficient to compute and can model loose and tight clothing alike, but might suffer from color ambiguities across similarly dressed subjects. Other, more complex identity tracking methods can be used instead when real-time performance is not needed, such as when processing pre-recorded sequences.

To be able to match subjects robustly, we assign current detections to previously known identities not only based on appearance similarity, $S^A_{i,k} = (A_i[t]-A_k[t-1])^2$, but also on the 2D pose similarity $S^{P2D}_{i,k}(i,k) = (P^{2D}_{i[t]}-P^{2D}_{k[t-1]})^2$ and 3D pose similarity $S^{P3D}_{i,k}(i,k) = (P^{3D}_{i[t]}-P^{3D}_{k[t-1]})^2$.
A threshold on the dissimilarity is set to detect occlusions, persons leaving the field of view, and new persons entering. That means the number of persons $K[t]$ can change.
Person identities are maintained for a certain number of frames after disappearance, to allow for re-identification after momentary occlusions such as those caused by the tracked subjects passing behind an occluder.
We update the appearance histogram of known subjects at arrival time and every $30$ seconds to account for appearance changes such as varying illumination. 
\subsection{Relative Bone Length and Absolute Height Calculation}
Relative bone length between body parts is a scale-invariant property that is readily estimated by $P^{3D}_k$ in \textit{Stage II}. To increase robustness, we take the normalized skeleton bone lengths $b_k$ as the distance between linked joints in $P^{3D}_k$ averaged across the first 10 frames.

Translating relative pose estimates from pixel coordinates to absolute 3D coordinates in cm is a difficult task as it requires either a reference object of known position and scale or knowledge of the person's height, which in turn can only be guessed with uncertainty from monocular footage~\cite{gunel2018face}. 

In Section~\ref{sec:kinematics} we explain how the camera relative position up to a scale is recovered through a re-projection constraint. We can optionally utilize the ground plane as reference geometry since camera calibration is less cumbersome than measuring the height of every person appearing in the scene. See the supplementary document for details.

\subsection{Kinematic Skeleton Fitting}
\label{sec:kinematics}

After 2D and 3D joint position prediction, we optimize for the skeletal pose $\{\theta_k[t]\}_{k=1}^{K[t]}$ of all $K[t]$ people in the scene, with $\theta_k[t] \in \mathbb{R}^{D}$ where $D=29$ is the number of degrees of freedom (DOF) for one skeleton. 
Both, per-frame 2D and 3D pose estimates from previous stages are temporally filtered~\cite{Casiez:2012} before skeleton fitting. Note that $\theta_k \in \mathbb{R}^D$ describes the pose of a person in terms of joint angles of a fixed skeleton plus the global root position, meaning our final output is directly compatible with CG character animation pipelines. 
We jointly fit to both 2D and root-relative 3D predictions as this leads to better reprojection error while maintaining plausible and robust 3D articulation.
We estimate $\theta_k[t]$ by minimizing the fitting energy
\begin{align}
\mathcal{E}(\theta_1[t], \cdots, \theta_K[t]) &= 
w_\text{3D}E_\text{3D} + 
w_\text{2D} E_\text{2D} + 
w_\text{lim} E_\text{lim} \nonumber\\
&+ w_\text{temp} E_\text{temp} + 
w_\text{depth} E_\text{depth}
\; .
\end{align}
We formulate $\frac{\partial \mathcal{E}}{\partial \theta_k[t]}$ in closed form to perform efficient minimization by gradient descent using a custom implementation. The influence of the individual terms is balanced with 
$w_\text{3D} = 9\mathrm{e}{-1}$,
$w_\text{2D} = 1\mathrm{e}{-5}$,
$w_\text{lim} = 5\mathrm{e}{-1}$,
$w_\text{temp} = 1\mathrm{e}{-7}$, and
$w_\text{depth} = 8\mathrm{e}{-6}$. In the following, we explain each term in more detail.
\paragraph{3D Inverse Kinematics Term:}
The 3D fitting term measures the 3D distance between predicted root-relative 3D joint positions $P^{3D}_k[t]$ and the root-relative joint positions in the skeleton $\bar{\mathcal{P}}(\theta_k[t],b_k)$ posed by forward kinematics for every person $k$, joint $j$ and previously estimated relative bone lengths $b_k$,
\begin{equation}
E_\text{3D} = \sum_{k=1}^K \sum_{j=1}^{J3D} || \bar{\mathcal{P}}(\theta_k[t],b_k)_{j} - P^{3D}_{k,j}[t] ||_2^2 \quad .
\end{equation}

\paragraph{2D Re-projection Term:}
The 2D fitting term is calculated as the 2D distance between predicted 2D joint positions $P^{2D}_k[t]$ and the projection of the skeleton joint positions $\mathcal{P}(\theta_k[t],b_k)_{j}$ for every person $k$ and joint $j$,
\begin{equation}
E_\text{2D} = \sum_{k=1}^K \sum_{j=1}^{J2D} w^{2D}_{j} c_{j,k} || \Pi (h_k \mathcal{P}(\theta_k[t],b_k))_{j} - P^{2D}_{k,j}[t]\,||_2^2 \quad ,
\end{equation}
where $c$ is the 2D prediction confidence, $w^{2D}_{j}$ is per-joint relative weighting, and $\Pi$ is the camera projection matrix. 
The lower limb joints have a relative weighting of $1.7$, elbows $1.5$ and wrist joints $2.0$ as compared to torso joints (hips, neck, shoulders).
Note that $\mathcal{P}$ outputs unit height, the scaling with $h_k$ maps it to metric coordinates, and the projection constraint thereby reconstructs absolute position in world coordinates.

\paragraph{Joint Angle Limit Term:}
The joint limits regularizer enforces a soft limit on the amount of joint angle rotation based on the anatomical joint rotation limits $\theta^{min}$ and $\theta^{max}$. We write it as 
\begin{equation}
E_\text{lim} = \sum_{k=1}^K \sum_{j=7}^{D}
\begin{cases}
(\theta_j^{min} - \theta_{k,j}[t])^2 &, \text{if } \theta_{k,j}[t] < \theta_j^{min} \\
(\theta_{k,j}[t] - \theta_j^{max})^2 &, \text{if } \theta_{k,j}[t] > \theta_j^{max} \\
0 &, \text{otherwise}
\end{cases}
\quad ,
\end{equation}
where we start from $j=7$ since we do not have limits on the global position and rotation parameters.
Note that our neural network is trained to estimate joint positions and hence has no explicit knowledge about joint angle limits.
Therefore, $E_\text{lim}$ ensures biomechanical plausibility of our results.

\paragraph{Temporal Smoothness Term:}
Since our neural network estimates poses on a per-frame basis, the results might exhibit temporal jitter.
The temporal stability of our estimated poses is improved by
\begin{equation}
E_\text{temp}(\Theta) = \sum_{k=1}^K || \nabla \theta_k[t-1] - \nabla \theta_k[t] ||_2^2 \quad ,
\end{equation}
where
the rate of change in parameter values, $\nabla \theta_k$, is approximated using backward differences. In addition, we 
penalize variations in the less constrained depth direction stronger, using the smoothness term $E_\text{depth} = || \theta_{k,2}[t]_z-\theta_{k,2}[t-1]_z ||$, where $\theta_{k,2}$ is the degree of freedom that drives the z-component of the root position.

\paragraph{Inverse Kinematics Tracking Initialization:}
For the first frame of a new person track, the local joint angles of the skeleton are fit to the 3D prediction only considering $E_{3D}$ and $E_{lim}$. The local angles are then held fixed while minimizing $E_{2D}$ to find the best fit for global translation and rotation of the skeleton. Subsequently, the complete energy formulation $\mathcal{E}(\theta_1[t], \cdots, \theta_K[t])$ is used.

\section{SelecSLS Net:\\ A Fast and Accurate Pose Inference CNN}
\label{sec:dlnas}
Our \textit{Stage I} core network is the most expensive component of our algorithm in terms of computation time. 
We evaluate various popular network architectures (see Table~1 in the supplemental document) on the task of single person 2D pose estimation, and determine that despite various parameter-efficient depthwise-convolution-based designs, for GPU-based deployment, ResNet architectures provide a better or comparable speed--accuracy tradeoff, and thus we use be them as \emph{baselines}.

ResNet-50~\shortcite{he_resnet_cvpr2016} 
 has been employed for other multi-person pose estimation methods such as Mehta\etal~\shortcite{mehta_3dv18}, Rogez \etal~\shortcite{rogez_lcrpp}, and Dabral \etal~\shortcite{dabral2019multi}. 
 %
Only on top-end hardware (Nvidia 1080TI, 11.4TFLOPs) does the ResNet-50 variant of our system run in real time. For more widespread lower performance hardware like ours (Nvidia 1080-MaxQ, 7.5TFLOPs), ResNet-50-based \textit{Stage I} forward pass takes $>30$ms, which, together with the additional overhead of $\approx15$ms from the other components of our system, does not reach real-time performance of $> 25$~fps.

 We therefore propose a new network architecture module, called \textit{SelecSLS} module, that uses short range and long range concatenation-skip connections in a selective way instead of additive-skip connections.  
SelectSLS is the main building block of the novel \textit{SelecSLS Net} architecture for the \textit{Stage I} core CNN. 
Additive-skip, as is used in ResNet architectures, performs element-wise addition to the features at the skip connection point, whereas concatenative-skip connections, as used in DenseNet, performs concatenation along the channel-dimension. 
Our new selective use of concatenation-skip connectivity promotes information flow through the network, without the exorbitant memory and compute cost of the full connectivity of DenseNet. 
Our new \textit{Stage I} network shows comparable accuracy to a ResNet-50 core with a substantially faster inference time ($\approx1.4\times$), across single person and multi-person 2D and 3D pose benchmarks. 

\begin{figure}[t]
  \includegraphics[width=1.0\linewidth]{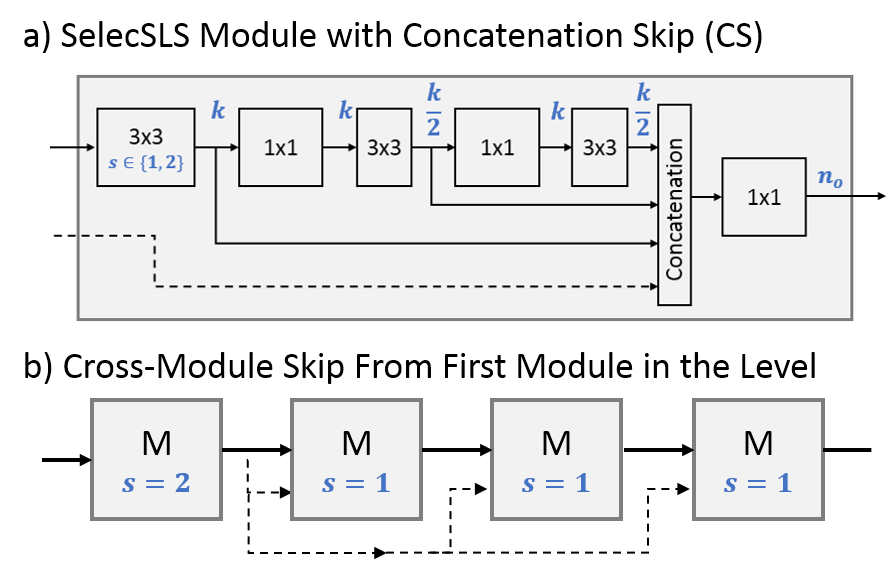}
  \caption
  {The proposed \textit{SelecSLS} module design (a) is comprised of interleaved $1\times1$ and $3\times3$ convolutions, and handles cross-module skip connections internally as concatenative-skip connections. The cross module skip connections themselves come from the first module that outputs features at a particular spatial resolution (b). See the supplemental document for ablation studies on alternative skip connectivity choices, through which this design emerged. Our design is parameterized by module stride ($s$), the number of intermediate features ($k$), and the number of module ouputs $n_o$.}
  \label{fig:network_module}
  \vspace{-0.3cm}
\end{figure}

\subsection{\textit{SelecSLS} Module}
Our \textit{Stage I} core network is comprised of building blocks, \textit{SelecSLS} modules, with intra-module short-range skip connectivity and cross-module longer-range skip connectivity.
The module design is as shown in Figure~\ref{fig:network_module} (a) and (b). It comprises of a series of $3\times3$ convolutions interleaved with $1\times1$ convolutions. This is to enable mixing of channels when grouped $3\times3$ convolutions are used. All convolutions are followed by batch normalization and ReLU non-linearity. The module hyperparameter $k$ dictates the number of features output by the convolution layers within the module. The outputs of all $3\times3$ convolutions ($2 k$) are concatenated and fed to a $1\times1$ convolution which produces $n_o$ features. The first $3\times3$ filters in the module are convolved with a stride of $1$ or $2$, which dictate the feature resolution of the entire module. 
The cross-module skip connection is the second input to the module. 

Different from the traditional inter-module skip connectivity pattern, which connects each module to the previous one, our design uses long range skip connectivity to the first module of each level (Figure~\ref{fig:network_module} (b)). This is intended to promote information flow through the network on the forward and the backward pass, and performs better than the traditional inter-module skip connectivity in practice. 
We define a level as all modules in succession which output feature maps of a particular spatial resolution. 
See the supplemental document for a detailed design ablation study leading to the proposed module design and the overall architecture.

\begin{table}[t]
\renewcommand{\tabcolsep}{2.0pt}
\centering
\caption{\textit{SelecSLS Net} Architecture: The table shows the network levels, overall number of modules, number of intermediate features $k$, the number of outputs of modules $n_o$, and the spatial resolution of features of the network proposed in Section~\ref{sec:dlnas}.}

\resizebox{0.9\columnwidth}{!}{%
\begin{tabular}{c|c|c|c|c|c|c}
\textbf{Level}  & \textbf{Output}     & \textbf{\textit{Module}}       & \textbf{Stride}       & \textbf{}  & \textbf{Cross-Module} & \multicolumn{1}{c}{} \\
 & \textbf{Resolution} & \textbf{Type} & \textbf{$s$} & \textbf{$k$} & \textbf{Skip Conn.} &  $n_o$\\ \hline

L0 & w/2 x h/2           & Conv. 3x3             & 2               & -          & -                   &                32            \\
\hline
 L1 & w/4 x h/4           & SelecSLS        & 2               & 64         & No                       & 64            \\
& w/4 x h/4           & SelecSLS       & 1               & 64         & First          &  \underline{128}          \\
\hline
L2 & w/8 x h/8           & SelecSLS        & 2               & 128        & No                              & 128           \\
& w/8 x h/8           & SelecSLS        & 1               & 128        & First                     & 128           \\
& w/8 x h/8           & SelecSLS        & 1               & 128        & First                  & \underline{288}           \\
\hline
L3 & w/16 x h/16         & SelecSLS        & 2               & 288        & No                               & 288           \\
& w/16 x h/16         & SelecSLS        & 1               & 288        & First                        & 288           \\
& w/16 x h/16         & SelecSLS        & 1               & 288        & First                      & 288           \\
& w/16 x h/16         & SelecSLS       & 1               & 288        & First                     & 416          
\end{tabular}
}

\label{tbl:net_arch}
\vspace{0.2cm}
\end{table}

\subsection{\textit{SelecSLS Net} Architecture}
\label{sec:dlnas_net}
Table~\ref{tbl:net_arch} shows the overall architecture of the proposed \textit{SelecSLS Net}, parameterized by the stride of the module ($s$), the intermediate features in the module ($k$), and number of outputs of the module ($n_o$). 
All $3\times3$ convolutions with more than $96$ outputs use a group size of $2$, and those with more than $192$ outputs use a group size of $4$.

\parahead{Design Evaluation}
We compare the proposed architecture against ResNet-50 and ResNet-34 architectures as core networks to establish appropriate baselines. We compare our proposed architecture's performance, both on 2D pose estimation and 3D pose estimation in Section \ref{sec:core_network_eval}. As we show in Tables~\ref{tbl:coco_eval_3d},~\ref{tbl:mpi_inf_jointwise}, and~\ref{tbl:mupots_comparison_compact}, the proposed architecture is $1.3-1.8\times$ faster than ResNet-50, while retaining the same accuracy.

\parahead{Memory Footprint}
The speed advantage of \emph{SelecSLS Net} over ResNet-50 stems primarily from its significantly smaller memory footprint. With a mini-batch comprised of a single $512\times320$px image, \emph{SelecSLS Net} occupies just 80\% of the memory (activations and bookkeeping for backward pass) occupied by ResNet-50. For larger mini-batch sizes, such as a mini-batch size of 32, the memory occupancy of \emph{SelecSLS Net} is 50\% of that of ResNet-50. Beyond the associated speed advantage, the significantly smaller memory footprint of \emph{SelecSLS Net} allows mini-batches of twice the size compared to ResNet-50, both for training and batched inference.   

\begin{table}[]
\renewcommand{\tabcolsep}{1.5pt}
\centering
\caption{Evaluation of 2D keypoint detections of the complete \textit{Stage I} of our system (both 2D and 3D branches trained), with different core networks on a subset of validation frames of MS COCO dataset. Also reported are the forward pass timings of the first stage on different GPUs (K80, TitanX (Pascal)) for an input image of size $512\times320$ pixels. We also show the 2D pose accuracy when using channel-dense supervision of $\{l_{j,k}\}_{j=1}^{J}$ in the 3D branch in place of our proposed channel-sparse supervision (Section~\ref{sec:pose_formulation}).}
\resizebox{\columnwidth}{!}{%
\begin{tabular}{l|cc|ccc|ccc}

\textbf{Core} & \multicolumn{2}{c|}{\textbf{FP Time}}          & \multicolumn{1}{l}{} & \multicolumn{1}{l}{} & \multicolumn{1}{l|}{} & \multicolumn{1}{l}{} & \multicolumn{1}{l}{} & \multicolumn{1}{l}{} \\
\textbf{Network}     & \textbf{K80} & \textbf{TitanX}  & \textbf{AP}          & \textbf{AP\textsubscript{0.5}}      & \textbf{AP\textsubscript{0.75}}      & \textbf{AR}          & \textbf{AR\textsubscript{0.5}}      & \textbf{AR\textsubscript{0.75}}     \\ \hline
ResNet-34                  & 29.0ms       & 6.5ms     & 45.0  & 72.0  & 46.1  & 49.9    & 74.4 & 51.6  \\
ResNet-50                  & 39.3ms       &10.5ms    & 46.6  & 73.0   & 48.9    & 51.4                 & 75.4               & 54.0               \\
\textit{SelecSLS}             & 28.6ms       & 7.4ms  & \textbf{47.0}   & \textbf{73.5}  & \textbf{49.5} & \textbf{51.8}  & 75.6  & \textbf{54.1}                 \\ \hline
\multicolumn{9}{l}{3D Branch With Channel-Dense $\{l_{j,k}\}_{j=1}^{J}$  Supervision} \\ \hline
\textit{SelecSLS}   & 28.6ms       & 7.4ms   & 46.8   & \textbf{73.5}  & 49.0 & 51.5  & \textbf{75.9}     & 53.8        \\
\end{tabular}}%
\label{tbl:coco_eval_3d}
\end{table}
\section{Results}
\label{sec:results}

In this section we evaluate the results of our real-time multi-person motion capture solution qualitatively and quantitatively on various benchmarks, provide extensive comparison with prior work, and conduct a detailed ablative analysis of the different components of our system. To ensure that the reported results on 3D pose benchmarks are actually indicative of the deployed system's performance, there is no test-time augmentation applied for our quantitative and qualitative results. We do not use procrustes alignment to the ground truth, nor do we rely on ground truth camera relative localization of the subjects to generate or modify our predictions.

For additional qualitative results, please refer to the accompanying video.

\subsection{System Characteristics and Applications}
First, we show that our system provides efficient and accurate 3D motion capture that is ready for live character animation and other interactive CG applications, rivaling depth-based solutions despite using only a single RGB video feed. 

\begin{figure}[t]
  \includegraphics[width=\linewidth, trim={0 19cm 0 0}, clip]{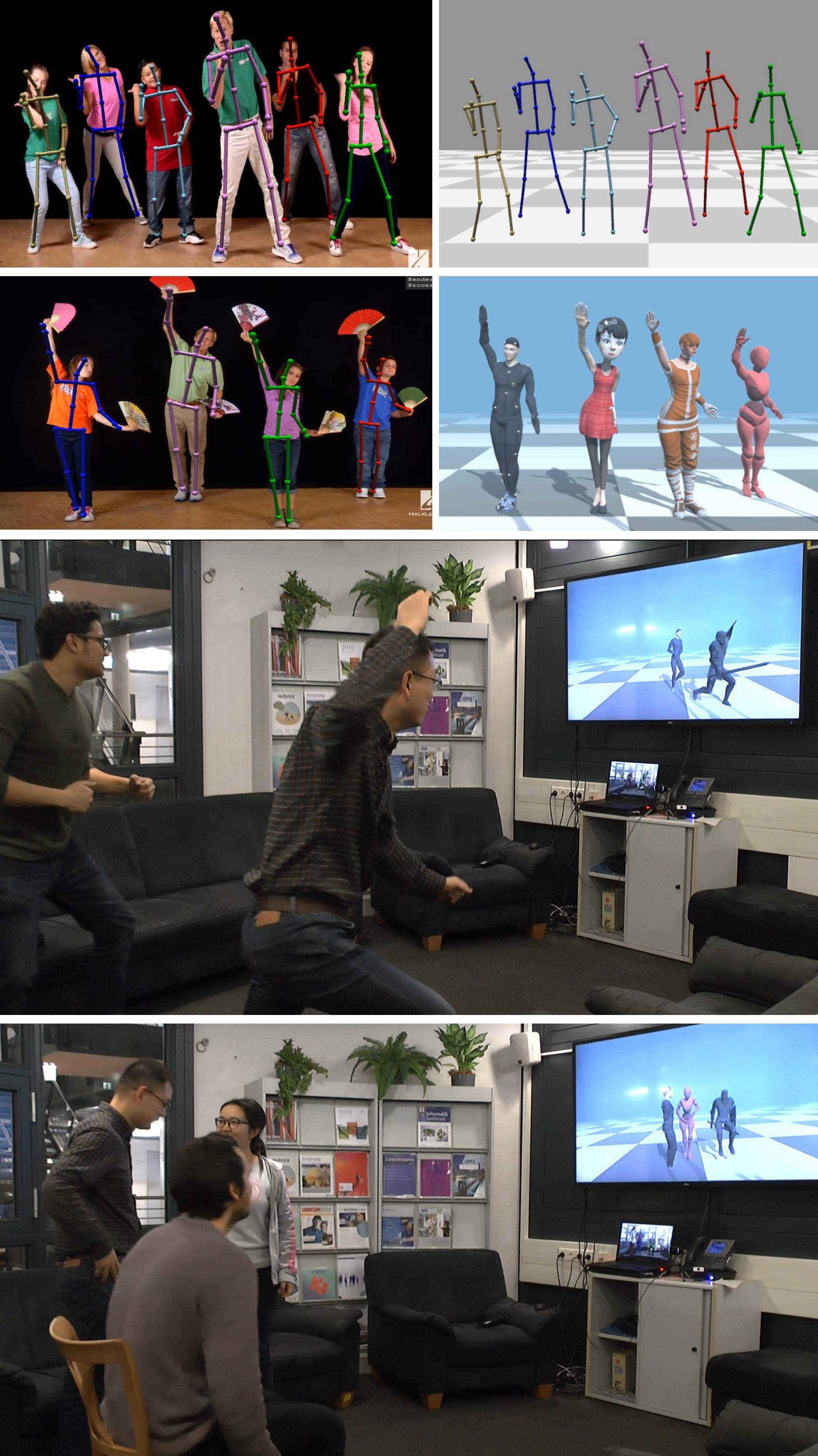}
  \caption
  {\textbf{Live Interaction and Virtual Character Control:} The temporally smooth joint angle predictions from \textit{Stage III} can be readily employed for driving virtual characters in real-time.
  The top two rows show our system driving virtual skeletons and characters with the motion captured in real time. On the bottom, our system is set up as a Kinect-like game controller, allowing subjects to interact with their virtual avatars live. Some images courtesy Music Express Magazine (\url{https://youtu.be/kX6xMYlEwLA}, \url{https://youtu.be/lv-h4WNnw0g}). See the accompanying video and the supplemental document for further character control examples. 
  }
  \label{fig:character_control}
\end{figure}

\paragraph{Real-time Performance:}
Our live system uses a standard webcam as input, and processes $512 \times 320$ pixel resolution input frames. For a scene with 10 subjects, the system running on a Desktop with an Intel Xeon E5 with 3.5~GHz and an Nvidia GTX 1080Ti is capable of processing input at $>30$~fps, while on a laptop with an Intel i7-8780H and a 1080-MaxQ it runs at $\approx27$~fps. 
On the laptop, \textit{Stage I} takes 21.5~ms, part association and feature extraction take 1~ms, \textit{Stage II} takes 1~ms, and \textit{Stage III} takes $\approx9$~ms (2.4~ms for identity matching to the previous frame, 6.8~ms for skeleton fitting).

We compare our timing against the faster but less accurate `demo' version of LCRNet++~\shortcite{rogez_lcrpp}, but compare our accuracy against the slower but more accurate version of LCRNet++.
LCRNet++ demo system uses ResNet-50 with less post-processing overhead than the accuracy-benchmarked system, and we measured its forward pass time on a TitanX-Pascal GPU (11 TFLOPs) to be 16~ms, while on a K80 GPU (4.1 TFLOPs) it takes $>$100~ms.
Our \textit{SelecSLS}-based system takes only 14~ms on TitanX-Pascal and 35~ms on a K80 GPU while producing more accurate per-frame joint position predictions (\textit{Stage II}) than the slower version of LCRNet++, as shown in Table~\ref{tbl:mupots_comparison}. An additional CPU-bound overhead of $\approx9$~ms for \textit{Stage III} results in temporally smooth joint-angle estimates which can readily be used to drive virtual characters.
The accompanying video contains examples of our live setup running on the laptop. 
Note that throughout the manuscript we report the timings of the various stages of our system on a set of GPUs with FP32 performance representative of current low-end and high-end consumer GPUs ($\approx4-12$TFLOPs).

\paragraph{Multi-Person Scenes and Occlusion Robustness:}
In Figure~\ref{fig:more_results}, we show qualitative results of our full system on several scenes containing multiple interacting and overlapping subjects, including frames from MuPoTS-3D~\shortcite{mehta_3dv18} and Panoptic~\shortcite{jooICCV2015} datasets. Single-person real-time approaches such as VNect~\shortcite{VNect_SIGGRAPH2017} are unable to handle occlusions or multiple people in close proximity as shown in Figure~5 in the supplemental document, while our approach can. Even offline single-person approaches such as HMMR \shortcite{humanMotionKanazawa19}, which utilize multi-person 2D parsing as a pre-processing step along with temporal information, fail under occlusions. Our approach is also more robust to occlusions than the multi-person approach of Mehta et al.~\shortcite{mehta_3dv18}, and unlike top-down approaches \citep{rogez_lcrpp,Moon_2019_ICCV_3DMPPE,dabral2019multi} it does not produce spurious predictions, as seen in Figure 5 in the supplemental document.
For further qualitative results on a variety of scene settings, including community videos and live scene setups, please refer to Figure~\ref{fig:more_results}, the accompanying video, and Figures~4 and 6 in the supplemental document.

\begin{figure}[t]
  \includegraphics[width=0.95\linewidth]{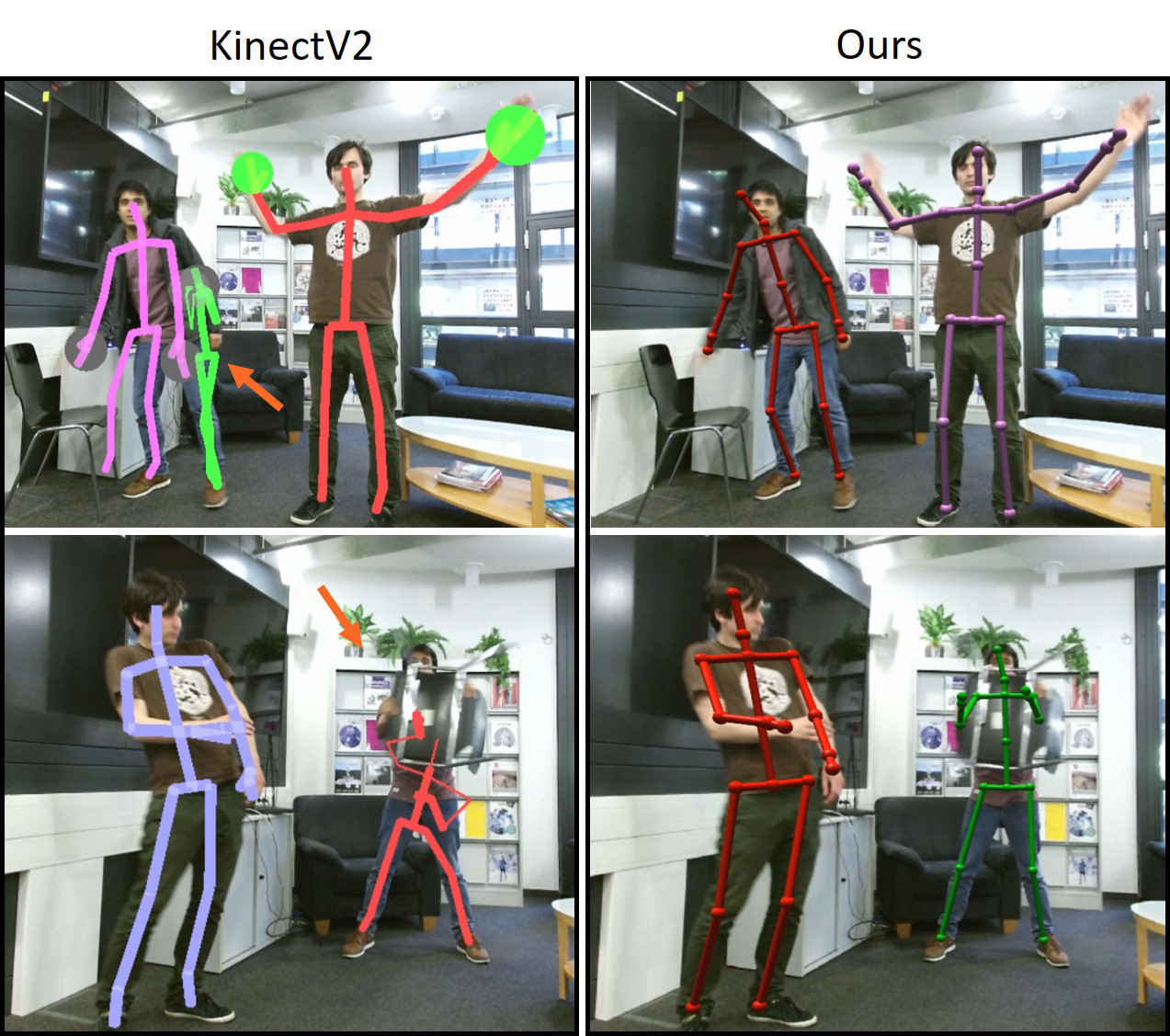}
  \caption
  {The quality of our pose estimates is comparable to depth sensing based approaches such as KinectV2, and our system handles certain cases of significant inter-personal overlap and cluttered scenes better than KinectV2. In the top row, due to scene clutter, KinectV2 predicts multiple skeletons for one subject. 
  In the bottom row, under occlusion, KinectV2 mispredicts the pose, while our approach succeeds.}
  \label{fig:kinect_comparisons}
  \vspace{-0.4cm}
\end{figure}

\begin{figure*}[]
  \includegraphics[width=1.0\linewidth]{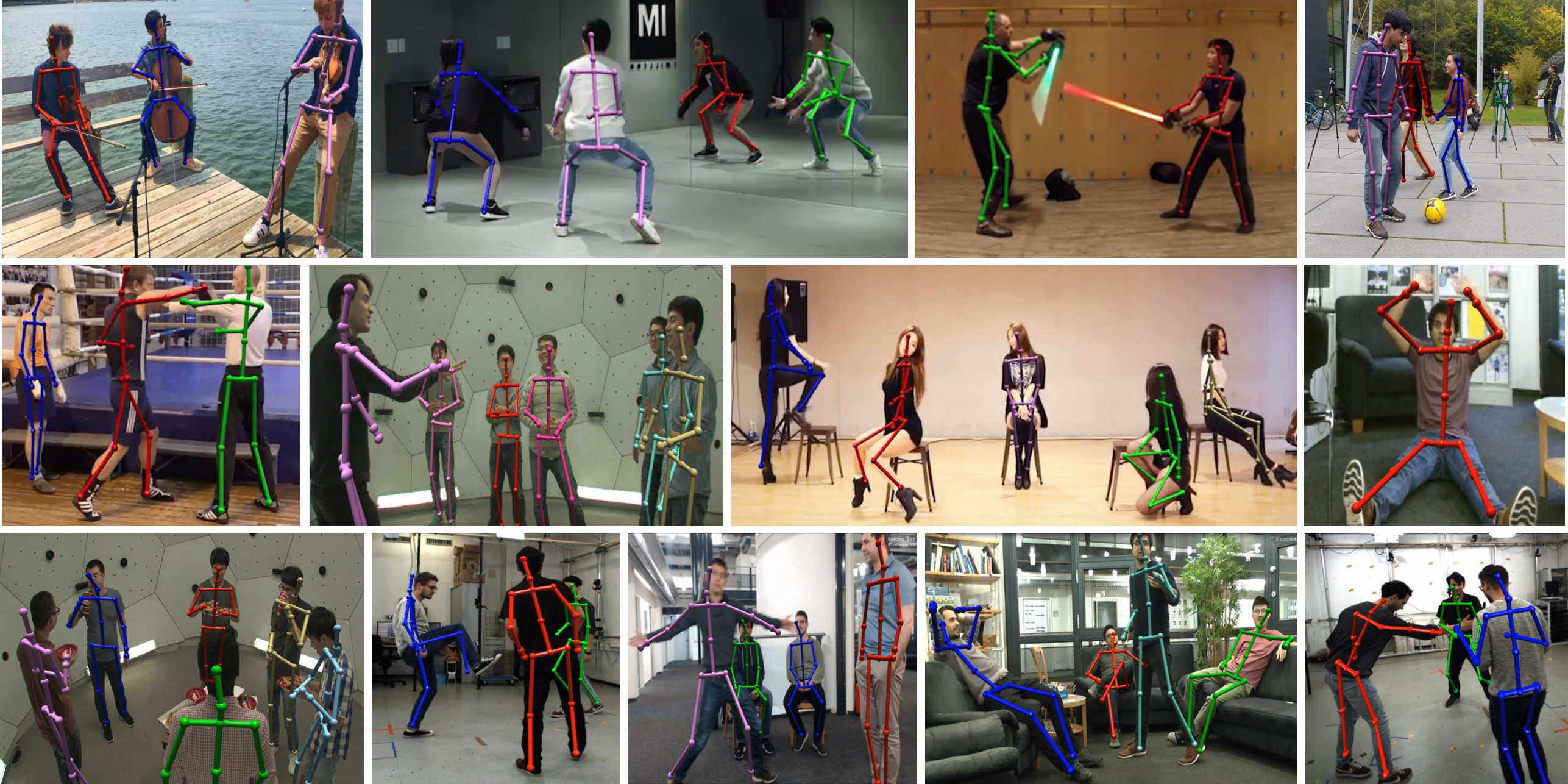}
  \caption
  {Monocular 3D motion capture results from our full system (\textit{Stage III}) on a wide variety of multi-person scenes, including Panoptic~\shortcite{joo_panoptic_iccv2015} and MuPoTS-3D~\shortcite{mehta_3dv18} datasets. Our approach handles challenging motions and poses, including interactions and cases of self-occlusion. 
  Some images courtesy KNG Music (\url{https://youtu.be/_xCKmEhKQl4}), 1MILLION TV (\url{https://youtu.be/9HkVnFpmXAw}), Boxing School Alexei Frolov (\url{https://youtu.be/dbuz9Q05bsM}), TPLA:Terra Prime Light Armory (\url{https://youtu.be/xmFVfUKr1MQ}), and Brave Entertainment (\url{https://youtu.be/ZhuDSdmby8k}).
  Please refer to the supplemental document and video for more results.}
  \label{fig:more_results}
\end{figure*}

\paragraph{Comparison With KinectV2:}
The quality of our pose estimates with a single RGB camera is comparable to off-the-shelf depth-sensing-based systems such as KinectV2 (Figure~\ref{fig:kinect_comparisons}), with our approach succeeding in certain cluttered scenarios where person identification from depth input would be ambiguous. The accompanying video contains further visual comparisons.

\paragraph{Character Animation:}
Since we reconstruct temporally coherent joint angles and our camera-relative subject localization estimates are stable, the output of our system can readily be employed to animate virtual avatars as shown in Figure~\ref{fig:character_control}. The video demonstrates the stability of the localization estimates of our system and contains further examples of real-time interactive character control with a single RGB camera.

\subsection{Performance on Single Person 3D Pose Datasets}
\label{res:single_person}
Our method is capable of real-time motion capture of multi-person scenes with notable occlusions. Previous single-person approaches, irrespective of runtime, would fail on this task. 
For completeness, we show that our method shows competitive accuracy on single-person 3D pose estimation. In Table~\ref{tbl:mpi_inf_comparison}, we compare the 3D pose output after \textit{Stage II} and \textit{Stage III} against other single-person methods on the MPI-INF-3DHP benchmark dataset~\cite{mehta_mono_3dv17} using the commonly used 3D Percentage of Correct Keypoints (\textbf{3DPCK}, higher is better), Area under the Curve (\textbf{AUC}, higher is better) and mean 3D joint position error (\textbf{MJPE}, lower is better). 
%

Similarly, with Stage II trained on Human3.6m (Table~\ref{tbl:h36m}), we again see that our system compares favourably to recent approaches designed to handle single-person and multi-person scenarios. 
Further, this is an example of the ability of our system to adapt to different datasets by simply retraining the inexpensive \textit{Stage II} network. 
Some approaches such as Sun \etal~\shortcite{sun2017compositional,sun2018integral}, obtain state-of-the-art accuracy on Human3.6m, but critically use the camera intrinsics and the ground-truth distance of the subject from the camera to convert their (u,v) predictions to (x,y), and also use flip augmentation at test time. Therefore, their reported results are not representative of the deployed system's performance.

\subsection{Performance on Multi-Person 3D Pose Datasets}
\label{res:multi_person}
We quantitatively evaluate our method's accuracy on the MuPoTS-3D monocular multi-person benchmark data set from Mehta \etal~\shortcite{mehta_3dv18}, which has ground-truth 3D pose annotations from a multi-view marker-less motion capture system.
We perform two types of comparison. In Table~\ref{tbl:mupots_comparison}(All), we compare on all annotated poses in sequences \textbf{T1-T20} of the annotated test set, including poses of humans that were not detected by our algorithm.
In Table~\ref{tbl:mupots_comparison}~(Matched), we compare only on annotated poses of humans detected by the respective algorithms. 

Both tables show that our per-frame predictions (\textit{Stage II}), as well as our temporally smooth model-fitting predictions (\textit{Stage III}) achieve better accuracy in terms of the 3D percentage of correct keypoints metric (\textbf{3DPCK}, higher is better) to LCRNet~\shortcite{rogez_lcr_cvpr17}, LCRNet++~\shortcite{rogez_lcrpp} and~\cite{mehta_3dv18}, while being comparable to other recent methods, namely~\cite{dabral2019multi}. The approach of~\citet{Moon_2019_ICCV_3DMPPE} exceeds the performance of our system, but it uses a prior person detection step, and passes resized person crops to the pose estimation network. As has been shown in prior work~\citep{cao_affinity_2017}, such an approach results in a higher accuracy, however, it cannot run at real-time frame rates, and the run time scales almost linearly with the number of subjects in the scene. In contrast, our approach runs in real-time for multi-person scenes, and only has a very mild dependence on the number of subjects in the scene.

Note that we apply no test-time augmentation or ensembling to our system, making the reported performance on various benchmarks accurately reflect the real per-frame prediction performance of our system.

In Table~\ref{tbl:mupots_comparison}~(Matched), we compare only on annotated poses of humans detected by the respective algorithms. 

\change{Additionally, we quantitatively evaluate our method's accuracy on the 3DPW~\cite{vonMarcard2018} monocular multi-person dataset, which comprises of videos recorded in in-the-wild settings with a phone camera. The 3D pose annotations for the dataset are generated from a combination of inertial-sensing measurements and video based approaches. We use the 24 sequences in the `test' split for evaluating. Our method does not use any data from 3DPW for training.
Unlike prior and concurrent work~\cite{humanMotionKanazawa19,kanazawa2018endtoend,kocabas2019vibe} which use bounding-box tracks of each subject, our approach handles all the subjects in the complete input frame jointly, and also matches person identities from frame to frame, all while running in real time. 
We report the mean 3D joint position error (MPJPE) in mm in Table~\ref{tbl:3dpw}. Our per-frame predictions (\textit{Stage II}) are comparable or better than most recent approaches, while being real-time and using no bounding box information. The person tracking heuristics for \textit{Stage III} are designed for stationary or slowly moving cameras, and the assumptions break under fast moving cameras in dense crowds where multiple subjects can have similar clothing and 2D poses. This leads to a noticeable decrease in accuracy with \textit{Stage III}, which can be improved by incorporating a better identity tracking approach, that is less prone to failure in dense crowds.
See the accompanying video for further results. 
}

\begin{table}[]
\centering
\caption{Results of Stage II predictions on Human3.6m, evaluated on all camera views of Subject 9 and 11 without alignment to GT. The Stage II network is trained with only Human3.6m. The top part has single person 3D pose methods, while the bottom part shows methods designed for multi-person pose estimation. Mean Per Joint Position Error (MPJPE) in millimeters is the metric used (lower is better). Note that our reported results do \emph{not use any test time augmentation}. Nor do we exploit ground truth 3D pose information in other ill-informed ways, such as rigid alignment to the ground truth or the use of ground truth depth from the camera to glean the image coordinate to 3D space transformation. }
\renewcommand{\tabcolsep}{2pt}

\begin{tabular}{l||c}
\multicolumn{1}{c||}{Method}             &  MPJPE (mm)                  \\ \hline 
\cite{katircioglu2018learning}   & 67.3\\
\cite{pavlakos_volumetric_cvpr17}  &  67.1    \\
\cite{zhou2017towards}  & 64.9   \\
\cite{martinez20173dbaseline}  & 62.9    \\
Hossain \& Little \shortcite{rayat2018exploiting}   & 58.3\\
\hline 
\cite{mehta_3dv18}        & 69.9          \\ 
LCRNet++\shortcite{rogez_lcrpp} (Res50)     & 63.5  \\ 
LCRNet+\shortcite{rogez_lcrpp} (VGG16)      & 61.2   \\
\cite{Moon_2019_ICCV_3DMPPE}    & 54.4    \\ 
Ours (\textit{Stage II})  &    63.6      \\
\end{tabular}
\vspace{-0.2cm}

\label{tbl:h36m}
\vspace{0.2cm}
\end{table}
\begin{table}[]
\renewcommand{\tabcolsep}{1.5pt}
\centering
\caption{Comparison on the single person MPI-INF-3DHP dataset. Top part are methods designed and trained for single-person capture. Bottom part are multi-person methods \emph{trained for multi-person capture} but evaluated on single-person capture.
Metrics used are: 3D percentage of correct keypoints ({3DPCK}, higher is better), area under the curve ({AUC}, higher is better) and mean 3D joint position error ({MJPE}, lower is better). * Indicates that \emph{no test time augmentation is employed}. \dag Indicates that \emph{no ground-truth bounding box information is used} and the complete image frame is processed.}
\begin{tabular}{l|ccc}
\textbf{Method}                  &          \textbf{3DPCK} & \textbf{AUC} & \textbf{MJPE} \\ \hline
\cite{VNect_SIGGRAPH2017}                & 78.1         & 42.0         & 119.2          \\
\cite{VNect_SIGGRAPH2017}*                & 75.0         & 39.2         & 132.8          \\
\cite{nibali20183d}                             & 87.6         & 48.8         & 87.6           \\
\cite{yang20183d}                     & 69.0         & 32.0         &  -          \\
\cite{zhou2017towards}                     & 69.2         & 32.5         &  -          \\
\cite{pavlakos2018ordinal}                     & 71.9         & 35.3         &  -          \\
\cite{dabral2018learning}                     & 72.3         & 34.8         & 116.3          \\
\cite{kanazawa2018endtoend}                     & 72.9         & 36.5         & 124.2 \\        
\cite{mehta_3dv18}                         & 76.2         & 38.3        & 120.5          \\
\hline
\cite{mehta_3dv18}                         & 74.1         & 36.7         & 125.1          \\
\cite{mehta_3dv18}*                         & 72.1         & 35.1         & 130.3          \\
\multicolumn{1}{l|}{Ours (\textit{Stage II})*\dag}         & {82.8}         & {45.3} & {98.4}          \\ 
\multicolumn{1}{l|}{Ours (\textit{Stage III})*\dag}         & {77.8}         & {38.9} & {115.0}          \\ 

\end{tabular}
\label{tbl:mpi_inf_comparison}
\end{table}
\begin{table}[]
\centering
\caption{Comparison of \textit{Stage II} prediction limb joint 3D pose accuracy on MPI-INF-3DHP (Single Person) for different core network choices with our channel-sparse supervision of 3D pose branch of \textit{Stage I}, as well as a comparison to channel-dense supervision. Metrics used are 3DPCK and AUC (higher is better).}

\resizebox{1.0\columnwidth}{!}{%
\begin{tabular}{l|c|c|c|c||c|c}
\multicolumn{1}{l|}{}     & \multicolumn{4}{c||}{\textbf{3DPCK}} & \multicolumn{2}{c}{\textbf{Total}} \\ \hline
\multicolumn{1}{l|}{}     & \textbf{Elbow} & \textbf{Wrist} & \textbf{Knee} & \textbf{Ankle} &  \textbf{3DPCK}&\textbf{AUC}  \\ \hline
ResNet-34                     & 79.6          & 61.2           & 83.0         & 52.7           & 79.3 & 41.8\\
ResNet-50                     & \textbf{82.4}           & {61.8}           & {87.1}          & 58.9          & 82.0   & 44.1         \\
\multicolumn{1}{l|}{SelecSLS}                  & 81.2           & \textbf{62.0}           & \textbf{87.6}          & \textbf{63.3}           & \textbf{82.8}  & \textbf{45.3}  \\ \hline
\multicolumn{7}{l}{Channel-Dense $ \{l_{j,k}\}_{j=1}^{J}$ Supervision \T }  \\ \hline
\multicolumn{1}{l|}{SelecSLS}                  & 79.0           & 60.2           & 82.5          & 59.0          & 80.1    &43.3 \\ 
\multicolumn{7}{l}{} \\

\end{tabular}
\vspace{-0.3cm}
} 
\label{tbl:mpi_inf_jointwise}
\end{table}
\begin{table}[]
\renewcommand{\tabcolsep}{1.5pt}
\centering
\caption{\change{Evaluation of 3D joint position error on the `test' split of 3DPW. The error is reported as the Mean Per-Joint Position Error (MPJPE) in mm for the following 14 joints: head, neck, shoulders, elbows, wrists, hips, knees, ankles. PA-MPJPE indicates the error after procrustes alignment. Lower is better.
Note that unlike prior work~\cite{humanMotionKanazawa19,kanazawa2018endtoend,kocabas2019vibe} which uses bounding-box crops around the subject and the supplied annotations to establish temporal association, our approach handles all the subjects in the complete input frame jointly, and also matches person identities from frame to frame, all while running in real time. The person identity tracking heuristics are designed for a stationary or slow moving camera, but the faster moving camera in 3DPW as well as the dense crowds cause identity tracking failures, which results in the markedly worse performance of \textit{Stage III} compared to \textit{Stage II}. 
}}

\begin{tabular}{l|c|c|}
  \multicolumn{1}{c|}{Method}           & ~~MPJPE ~~ &  ~PA-MPJPE~ \\ \hline
\cite{sun2019human}            &   -    & 69.5     \\
\cite{arnab2019exploiting} &        -      & 72.2     \\
\cite{kolotouros2019convolutional}          &     -         & 70.2     \\
\cite{kanazawa2018endtoend}          & 130.0          & 76.7     \\
\cite{kolotouros2019learning}         & 96.9         & 59.2     \\
\cite{humanMotionKanazawa19}         & 116.5        & 72.6     \\
\cite{kocabas2019vibe}         & 93.5         & 56.5     \\ \hline
Ours(\textit{Stage II})     & 103.3   & 60.7     \\
Ours(\textit{Stage III})    & 134.2   & 80.3     
\end{tabular}
\label{tbl:3dpw}
\end{table}

\begin{table*}[t]
\renewcommand{\tabcolsep}{1.5pt}
\centering
\caption{Comparison of our per-frame estimates (Stage II) on the MuPoTS-3D benchmark data set~\cite{mehta_3dv18}. The metric used is 3D percentage of correct keypoints (\textbf{3DPCK}), so higher is better. The data set contains 20 test scenes \textbf{TS1-TS20}. We evaluate once on all annotated poses (top row - \textbf{All}), and once only on the annotated poses detected by the respective algorithm (bottom row - \textbf{Matched}). Our approach achieves comparable accuracy to the previous monocular multi-person 3D methods (SingleShot~\cite{mehta_3dv18}, LCRNet~\cite{rogez_lcr_cvpr17}, MP3D~\cite{dabral2019multi}, LCRNet++~\cite{rogez_lcrpp}) while having a drastically faster runtime. The accuracy of our system lags behind that of Root+PoseNet~\cite{Moon_2019_ICCV_3DMPPE} which uses a prior person detection step, runs offline, has its per-frame inference time scale linearly with the number of subjects in the scene. * Indicates \textbf{no} test time augmentation is used.}
\begin{tabular}{lccccccccccccccccccccc}
\multicolumn{1}{l|}{\textbf{All}}         & \textbf{TS1} & \textbf{TS2} & \textbf{TS3} & \textbf{TS4} & \textbf{TS5} & \textbf{TS6} & \textbf{TS7} & \textbf{TS8} & \textbf{TS9} & \textbf{TS10} & \textbf{TS11} & \textbf{TS12} & \textbf{TS13} & \textbf{TS14} & \textbf{TS15} & \textbf{TS16} & \textbf{TS17} & \textbf{TS18} & \textbf{TS19} & \multicolumn{1}{c|}{\textbf{TS20}} & \textbf{Total} \\ \hline
\multicolumn{1}{l|}{LCRNet*}      & 67.7         & 49.8         & 53.4         & 59.1         & 67.5         & 22.8         & 43.7         & 49.9         & 31.1         & 78.1          & 33.4          & 33.5          & 51.6          & 49.3          & 56.2          & 66.5          & 65.2          & 62.9          & 66.1          & \multicolumn{1}{c|}{59.1}          & 53.8           \\
\multicolumn{1}{l|}{Single Shot} & 81.0         & 59.9         & 64.4         & 62.8         & 68.0         & 30.3         & 65.0         & 59.2         & 64.1         & 83.9          & 67.2          & 68.3          & 60.6          & 56.5          & 69.9          & 79.4          & 79.6          & 66.1          & 64.3          & \multicolumn{1}{c|}{63.5}          & 65.0           \\
\multicolumn{1}{l|}{MP3D*}    & 85.1         & {67.9}         & {73.5}         & {76.2}         & 74.9         & {52.5}         & 65.7         & {63.6}         & 56.3         & 77.8          & {76.4}          & 70.1          & 65.3          & 51.7            & 69.5          & {87.0}          & 82.1          & {80.3}          & {78.5}         & \multicolumn{1}{c|}{{70.7}}            & {71.3}           \\ 
\multicolumn{1}{l|}{Root+PoseNet*}    & \textbf{94.4}         & \textbf{77.5}         & \textbf{79.0}         & \textbf{81.9}         & \textbf{85.3}         & \textbf{72.8}         & \textbf{81.9}         & \textbf{75.7}         & \textbf{90.2}         & \textbf{90.4}          & \textbf{79.2}          & \textbf{79.9}          & \textbf{75.1}          & \textbf{72.7}            & \textbf{81.1}          & \textbf{89.9}          & \textbf{89.6}          & \textbf{81.8}          & \textbf{81.7}          & \multicolumn{1}{c|}{\textbf{76.2}}            & \textbf{81.8}           \\ 
\multicolumn{1}{l|}{LCRNet++ (Res50)*}    & 87.3         & 61.9         & 67.9         & 74.6         & {78.8}         & 48.9         & 58.3         & 59.7         & {78.1}         & {89.5}          & 69.2          & {73.8}          & {66.2}          & 56.0            & {74.1}          & {82.1}          & 78.1          & {72.6}          & 73.1          & \multicolumn{1}{c|}{61.0}            & {68.9}           \\ \hline

\multicolumn{1}{l|}{{Ours (\textit{Stage II})}*}    & 89.7 & 65.4 & 67.8 & 73.3 & 77.4 & 47.8 & 67.4 & 63.1 & 78.1 & 85.1 & 75.6 & 73.1 & 65.0 & 59.2 & 74.1 & 84.6 & 87.8 & 73.0 & 78.1 & \multicolumn{1}{c|}{{71.2}}          & 72.1           \\
\multicolumn{1}{l|}{{Ours (\textit{Stage III})}*}  &  86.3 & 63.5 & 66.1 & 71.1 & 69.7 & 48.4 & {68.4} & 62.9 & 76.4 & 85.4 & 72.7 & {75.1} & 61.9 & {62.9} & 70.3 & 84.4 & 84.6 & 72.2 & 70.4        & \multicolumn{1}{c|}{{69.4}}          & 70.4           \\
                                 &              &              &              &              &              &              &              &              &              &               &               &               &               &               &               &               &               &               &               &                                    &                \\
\multicolumn{1}{l|}{\textbf{Matched}}     & \textbf{TS1} & \textbf{TS2} & \textbf{TS3} & \textbf{TS4} & \textbf{TS5} & \textbf{TS6} & \textbf{TS7} & \textbf{TS8} & \textbf{TS9} & \textbf{TS10} & \textbf{TS11} & \textbf{TS12} & \textbf{TS13} & \textbf{TS14} & \textbf{TS15} & \textbf{TS16} & \textbf{TS17} & \textbf{TS18} & \textbf{TS19} & \multicolumn{1}{c|}{\textbf{TS20}} & \textbf{Total}   \\ \hline
\multicolumn{1}{l|}{LCRNet*}      & 69.1         & 67.3         & 54.6         & 61.7         & 74.5         & 25.2         & 48.4         & 63.3         & 69           & 78.1          & 53.8          & 52.2          & 60.5          & 60.9          & 59.1          & 70.5          & 76            & 70            & 77.1          & \multicolumn{1}{c|}{81.4}          & 62.4           \\
\multicolumn{1}{l|}{Single Shot} & 81.0           & 64.3         & 64.6         & 63.7         & 73.8         & 30.3         & 65.1         & 60.7         & 64.1         & 83.9          & 71.5          & 69.6          & 69            & 69.6          & 71.1          & 82.9          & 79.6          & 72.2          & 76.2          & \multicolumn{1}{c|}{85.9}          & 69.8           \\
\multicolumn{1}{l|}{MP3D*}    & 85.8           & {73.6}         & 61.1         & {55.7}         & 77.9         & {53.3}         & {75.1}         & 65.5         & 54.2         & 81.3          & \textbf{82.2}          & {71.0}          & {70.1}          & 67.7          & {69.9}          & {90.5}          & {85.7}          & \textbf{86.3}          & 85.0          & \multicolumn{1}{c|}{{91.4}}          & 74.2             \\ 
\multicolumn{1}{l|}{Root+PoseNet*}    & \textbf{94.4}         & \textbf{78.6}         & \textbf{79.0}         & \textbf{82.1}         & {\textbf{86.6}}         & \textbf{72.8}         & \textbf{81.9}         & \textbf{75.8}         & \textbf{90.2}         & \textbf{90.4}          & 79.4          & \textbf{79.9}          & {\textbf{75.3}}          & 81.0            & \textbf{81.0}          & \textbf{{90.7}}          & \textbf{89.6}          & {83.1}          & 81.7          & \multicolumn{1}{c|}{77.2}            & \textbf{82.5}           \\ 
\multicolumn{1}{l|}{LCRNet++ (Res50)*}    & {88.0}           & {73.3}         & {67.9}         & {74.6}         & 81.8         & {50.1}         & 60.6         & 60.8         & {78.2}         & {89.5}          & 70.8          & {74.4}          & {72.8}          & 64.5          & {74.2}          & {84.9}          & {85.2}          & {78.4}          & 75.8          & \multicolumn{1}{c|}{74.4}          & -             \\ \hline

\multicolumn{1}{l|}{{Ours (\textit{Stage II})}*}    & 89.7 & 78.6 & 68.4 & 74.3 & 83.7 & 47.9 & 67.4 & 75.4 & 78.1 & 85.1 & 78.7 & 73.8 & 73.9 & 77.9 & 74.8 & 87.1 & 88.3 & 79.5 & \textbf{88.3} & \multicolumn{1}{c|}{{\textbf{97.5}}}          & 78.0      \\
\multicolumn{1}{l|}{{Ours (\textit{Stage III})}*}  &  86.3 & {76.2} & 66.7 & 72.1 & 75.3 & 48.5 & 68.4 & {75.1} & 76.4 & 85.4 & 75.7 & {75.8} & 70.3 & \textbf{82.9} & 70.9 & 86.9 & 85.1 & 78.7 & 79.5 & \multicolumn{1}{c|}{{95.0}}          & {76.2}           \\
\end{tabular}
\label{tbl:mupots_comparison}
\end{table*}

\begin{table}[]
\renewcommand{\tabcolsep}{2.0pt}
\centering
\caption{
Evaluations on the multi-person 3D pose benchmark MuPoTS-3D~\shortcite{mehta_3dv18} of \textit{Stage II} predictions for different \textit{Stage I} core network choices with channel-sparse supervision of 3D pose branch of \textit{Stage I}, as well as a comparison to channel-dense supervision. We evaluate on all annotated subjects using the 3D percentage of correct keypoints (\textbf{3DPCK}) metric, and also show the 3DPCK only for predictions that were matched to an annotation. 
}

\resizebox{0.75\columnwidth}{!}{%
\begin{tabular}{r|c|c|c}
    & \multicolumn{2}{c|}{\textbf{3DPCK}}    & \multicolumn{1}{c}{\textbf{\% Subjects}}             \\
\textbf{}       & \textbf{\qquad All \qquad} & \textbf{~Matched~}  & \textbf{Matched}\\ \hline
~~~ResNet-34~~~\T        &  69.3      & 75.2             &  92.1       \\
~~~ResNet-50~~~\T        & 71.8         & 77.2             &  \textbf{93.0}    \\
~~~SelecSLS~~~\T        & \textbf{72.1}         & \textbf{78.0}  & 92.5     \\ \hline
\multicolumn{4}{l}{Channel-Dense $ \{l_{j,k}\}_{j=1}^{J}$ Supervision}\T     
\\ \hline
~~~SelecSLS~~~\T        & 70.2         & 75.7              & 92.7       \\
\end{tabular}
} 
\label{tbl:mupots_comparison_compact}
\end{table}

\subsection{Core Network Architecture Evaluation}
\label{sec:core_network_eval}
We evaluate the performance of the proposed network architecture against ResNet baselines on 2D pose estimation and 3D pose estimation tasks.
For ResNet, we keep the network until the first residual module in level-5 and remove striding from level-5.
For 2D pose estimation, we evaluate on a held-out 1000 frame subset of the MS-COCO validation set, and report the Average Precision (AP) and Recall (AR), as well as inference time on different hardware in Table~\ref{tbl:coco_eval_3d}. The performance of SelecSLS Net (47.0 AP, 51.8 AR) is better than ResNet-50 (46.6 AP, 51.4 AR), while being $1.3-1.4\times$ faster on GPUs. The inference speed improvement increases to $1.8\times$ when running on a CPU, as shown in the supplemental document. 

We evaluate our network trained for multi-person pose estimation (after Stage II) on MPI-INF-3DHP~\cite{mehta_mono_3dv17} single person 3D pose benchmark, comparing the performance of different core network architectures.
In Table~\ref{tbl:mpi_inf_jointwise}, we see that using our \textit{SelecSLS} core architecture we overall perform significantly better than ResNet-34 and slightly better than ResNet-50, with a higher 3DPCK and AUC and a lower MPJPE error. \textit{SelecSLS} particularly results in significantly better performance for lower body joints (Knee, Ankle) than the ResNet baselines.

Similarly, our \textit{SelectSLS} network architecture outperforms ResNet-50 and ResNet-34 on the multi person 3D pose benchmark MuPoTS-3D, as shown in Table~\ref{tbl:mupots_comparison_compact}. 

With \textit{Stage II} trained on the single-person pose estimation task on Human3.6m, we again see that our proposed faster core network architecture outperforms ResNet baselines. The use of \textit{SelecSLS} results in a mean per joint position error of 63.6mm, compared to 64.8mm using ResNet-50 and 67.6mm using ResNet-34.

\begin{table}[]
\renewcommand{\tabcolsep}{1.5pt}
\centering
\caption{Comparison of limb joint 3D pose accuracy on MuPoTS-3D (Multi Person) for predictions from \textit{Stage II} and \textit{Stage III} of our proposed design.
The metric used is 3D Percentage of Correct Keypoints (3DPCK), evaluated with a threshold of 150mm.}
\vspace{-0.2cm}
\begin{tabular}{l|c|c|c|c|c||c|c|c}

\multicolumn{6}{l||}{}     &  \multicolumn{3}{c}{\textbf{All Joints}} \\ \hline
\multicolumn{1}{l|}{}     & \textbf{Shoul} &\textbf{Elbow} & \textbf{Wrist} & \textbf{Knee} & \textbf{Ankle} & \textbf{Total} &\textbf{Vis.}&\textbf{Occ.}  \\ \hline
\textit{Stage II}        &\textbf{81.4}          & 65.8           & 53.2           & 71.0         & 47.3           & \textbf{72.1}  &\textbf{76.0} &\textbf{58.8} \\ 
\textit{Stage III}       &{73.8}           & \textbf{67.2}           & \textbf{54.0}           & \textbf{75.8}          & \textbf{54.3}          & {70.4}  &{75.0} &55.2 \\ \hline
\end{tabular}
\label{tbl:mupots_jointwise_comparison}
\vspace{-0.3cm}
\end{table}
\subsection{Channel-Sparse 3D Pose Encoding Evaluation}
\label{sec:sparse_v_dense}
As discussed at length in Section~\ref{sec:pose_formulation}, different choices for supervision of $ \{l_{j,k}\}_{j=1}^{J}$ have different implications. Here we show that our channel-sparse supervision of the encoding, such that only the local kinematic context is accounted for, performs better than the nai\"ve channel-dense supervision. 

The 2D pose accuracy of the \textit{Stage I} network with channel-dense supervision of 3D branch is comparable to that with channel-sparse supervision, as shown in Table~\ref{tbl:coco_eval_3d}. However, our proposed encoding performs much better across single person and multi-person 3D pose benchmarks. 

Table~\ref{tbl:mpi_inf_jointwise} shows that channel-sparse encoding significantly outperforms channel-dense encoding, yielding an overall 3DPCK of 82.8 compared to 80.1 3DPCK for the latter. As shown in Table~4 in the supplemental document, the difference particularly emerges for difficult pose classes such as sitting on a chair or on the floor, where our channel-sparse supervision shows substantial gains. 
Also refer to the supplemental document for additional ablative analysis of input to \textit{Stage II}.

\subsection{Evaluation of Skeleton Fitting (Stage III)}
Skeleton fitting to reconcile 2D and 3D poses across time results in smooth joint angle estimates which can be used to drive virtual characters. The accompanying video shows that the resulting multi-person pose estimates from our real-time system are similar or better in quality than the recently published offline single-person approach of \cite{humanMotionKanazawa19}, while also succeeding under significant occlusions.

On the multi-person benchmark MuPoTS-3D in Table~\ref{tbl:mupots_comparison}, the overall performance of \emph{Stage III} appears to lag behind that of \emph{Stage II}.
However, the efficacy of \emph{Stage III} is revealed through joint-wise breakdown of performance on MuPoTS-3D in Table~\ref{tbl:mupots_jointwise_comparison}. \emph{Stage III} predictions show a significant improvement for limb joints such as elbows, wrists, knees, and ankles over \emph{Stage II}. The decrease in accuracy for some other joints, such as the head, is partly due to lack of 2D constraints imposed for those joints in our system. 
The overall breakdown of accuracy by joint visibility shows that \emph{Stage III} improves the accuracy for visible joints, i.e. where 2D constraints were available, while slightly decreasing accuracy for occluded joints.

On the single-person benchmark MPI-INF-3DHP (Table~\ref{tbl:mpi_inf_comparison}), the performance of \emph{Stage III} is comparable to other methods specifically designed for single-person scenarios, however it loses accuracy compared to \emph{Stage II} predictions for similar reasons as in the multi-person case. 
For complex poses, not all 2D keypoints may be detected. This does not pose a problem for \textit{Stage II} predictions due to redundancies in the pose encoding, however, analogous to the case of occluded joints discussed before, missing 2D keypoints can cause the \textit{Stage III} accuracy to worsen.

However, despite a minor decrease in quantitative accuracy due to \textit{Stage III}, it produces a marked improvement in the quality and temporal stability of the pose estimates, and increases the accuracy of the end effectors. The resulting, temporally smooth joint angle estimates can be used in interactive graphics applications. Refer to the accompanying video for qualitative results, and virtual character control examples.
%

\section{Discussion and Future Work}
\label{sec:discussion}

Our approach is the first to perform real-time 3D motion capture of challenging multi-person scenes with one color camera. Nonetheless, it has certain limitations that will be addressed in future work. 

As with other monocular approaches, the accuracy of our method is not comparable yet to the accuracy of multi-view capture algorithms. 
Failure cases in our system can arise from each of the constituent stages. The 3D pose estimates can be incorrect if the underlying 2D pose estimates or part associations are incorrect. Also, since we require the neck to be visible for a successful detection of a person, scenarios where the neck is occluded result in the person not being detected despite being mostly visible. See Figure~3 in the supplemental document for a visual representation of the limitations. 

Our algorithm successfully captures the pose of occluded subjects even under difficult inter-person occlusions that are generally hard for monocular methods. However, the approach still falls short of reliably capturing extremely close interactions, like hugging. Incorporation of physics-based motion constraints could further improve pose stability in such cases, may add further temporal stability, and may allow capturing of fine-grained interactions of persons and objects.

In some cases individual poses have higher errors for a few frames, \eg after strong occlusions (see accompanying video for example). However, our method manages to recover from this.   
The kinematic fitting stage may suffer from inaccuracies under cases of significant inter-personal or self occlusion, making the camera relative localization less stable in those scenarios.
Still, reconstruction accuracy and stability 
is appropriate for many real-time applications.

Our algorithm is fast, but the relatively simple identity tracker may swap identities of people when tracking through extended person occlusions, drastic appearance changes, and similar clothing appearance. More sophisticated space-time tracking would be needed to remedy this. 
As with all learning-based pose estimation approaches, pose estimation accuracy worsens on poses very dissimilar from the training poses. To approach this, we plan to extend our algorithm in future such that it can be refined in an unsupervised or semi-supervised way on unlabeled multi-person videos.

Our \textit{SelecSLS Net} leads to a drastic performance boost. 
There are other strategies that could be explored to further boost the speed of our network and convolutional architectures in general, or target it to specific hardware, such as using depthwise $3\times3$ convolutions or factorized $3\times3$ convolutions~\cite{romera2018erfnet,szegedy2017inception} or binarized operations~\cite{bulat2017binarized}, all of which our proposed design can support. Its faster inference speed and significantly smaller memory footprint than ResNet-50 without compromising accuracy makes it an attractive candidate to replace ResNet core networks for a broad range of tasks beyond body pose estimation. It can support much larger training and inference batch sizes, or can be trained on lower-end GPUs with similar batch sizes as ResNet-50 on higher-end GPUs.

\section{Conclusion}
\label{sec:conclusion}
We present the first real-time approach for multi-person 3D motion capture using a single RGB camera. It operates in generic scenes and is robust to occlusions both by other people and objects. It provides joint angle estimates and localizes subjects relative to the camera. To this end, we jointly designed pose representations, network architectures, and a model-based pose fitting solution, to enable real-time performance. One of the key components of our system is a new CNN architecture that uses selective long and short range skip connections to improve the information flow and have a significantly smaller memory footprint, allowing for a drastically faster network without compromising accuracy. The proposed system runs on consumer hardware at more than $30$~fps while achieving state-of-the-art accuracy. We demonstrate these advances on a range of challenging real-world scenes.

\clearpage
 \begin{center}
 \textbf{\large Supplemental Document:\\XNect: Real-time Multi-Person 3D Motion Capture with a Single RGB Camera}
 \end{center}
 \setcounter{equation}{0}
 \setcounter{figure}{0}
 \setcounter{table}{0}
 \setcounter{section}{0}

Here we present additional qualitative results of our approach, ablation studies of design variations of SelecSLS architecture, further ablation studies of our proposed pose representation, and additional details about \textit{Stage III} of our system. Refer to the main manuscript, the accompanying video, and the project website (\url{http://gvv.mpi-inf.mpg.de/projects/XNect/}) for further details and results.

\section{\textit{SelecSLS Net} Design Evaluation}
\label{sec:dlnas_net}
\begin{figure*}[t]
  \includegraphics[width=1.0\linewidth]{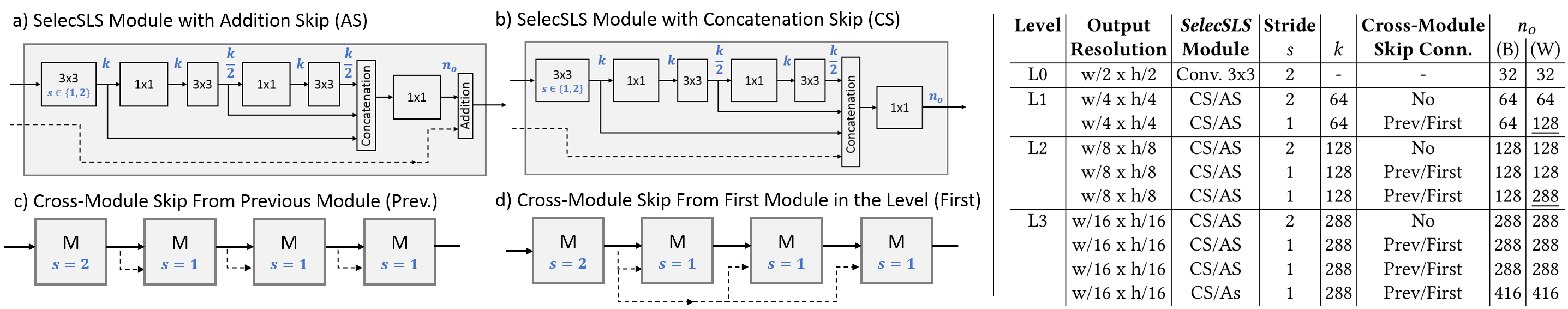}
  \caption
  {Variants of \textit{SelecSLS} module design (a) and (b). Both share a common design comprised of interleaved $1\times1$ and $3\times3$ convolutions, with different ways of handling cross-module skip connections internally: (a) as additive-skip connections, or (b) as concatenative-skip connections. The cross module skip connections can themselves come either from the previous module (c) or from the first module which outputs features at a particular spatial resolution (d). In addition to the different skip connectivity choices, our design is parameterized by module stride ($s$), the number of intermediate features ($k$), and the number of module ouputs $n_o$.
 The table on the right shows the network levels, overall number of modules, number of intermediate features $k$, and the spatial resolution of features of the network designs we evaluate in Section~\ref{sec:dlnas_net}. The design choices evaluated are the type of module (additive skip \textit{AS} vs concatenation skip \textit{CS}), the type of cross module skip connectivity (From previous module (\textit{Prev}) or first module (\textit{First} in the level), and the scheme for the number of outputs of modules $n_o$ ((B)ase or (W)ide). 
  }
  \label{fig:network_module_full}
\end{figure*}

Figure~\ref{fig:network_module_full} shows variants of the overall architecture of the proposed \textit{SelecSLS Net} that were considered. The architecture is parameterized by the type of module (\textit{SelecSLS} concatenation-skip \textit{CS} vs addition-skip \textit{AS}), the stride of the module ($s$), the intermediate features in the module ($k$), cross- module skip connectivity (previous module or first module in the level), and number of outputs of the module ($n_o$ (B)ase case). With the aim to promote information flow in the network, we also consider (W)ider $n_o$ at transitions in spatial resolution. All $3\times3$ convolutions with more than $96$ outputs use a group size of $2$, and those with more than $192$ outputs use a group size of $4$.

We experimentally determine the best network design by testing the \textit{Stage I} network with a \textit{SelecSLS Net} core on 2D multi-person pose estimation, \ie only using the 2D branch, which plays an integral role in the overall pipeline. Our conclusions transfer to the full \textit{Stage I} network, as further evidenced in Section 7.4 in the main manuscript.

As mentioned in Section 6.2 in the main manuscript, and shown in Table~\ref{tbl:lsp_eval}, for GPU-based deployment ResNet architectures provide a better or comparable speed--accuracy tradeoff to various parameter-efficient depthwise-convolution based designs.
Thus, we compare against ResNet-50 and ResNet-34 architectures as core networks to establish the appropriate baselines. For ResNet, we keep the network until the first residual module in level-5 and remove striding from level-5.
We evaluate on a held-out 1000 frame subset of the MS-COCO validation set, and report the Average Precision (AP) and Recall (AR), as well as inference time on different hardware in Table~\ref{tbl:coco_eval}. Using the \textit{AS} module with \textit{Prev} connectivity  
and $n_o$(B) outputs for modules, the performance as well as the inference time on an Nvidia K80 GPU is close to that of ResNet-34. Using \textit{CS} instead of addition-skip significantly improves the average precision from 47.0 to 47.6, and the average recall from 51.7 to 52.6. Switching the number of module outputs to the wider $n_o$(W) scheme leads to further improvement in AP and AR, at a slight increase in inference time. Using 
\textit{First} connectivity 
further improves performance, namely to 48.6 AP and 53.3 AR,
reaching close to ResNet-50 in AP (48.8) and performing slightly better with regard to AR (53.2).
Still our new design has a 1.4-1.8$\times$ faster inference time across all devices. 
We also evaluate the publicly available model of ~\cite{cao_affinity_2017} on the same validation subset. Their multi-stage network is $11$ percentage points better on AP and AR than our network, at the price of being $10-20\times$ slower. 
The follow-up versions \cite{cao2018openpose} are $\approx2\times$ faster on the GPU  and $\approx5\times$ slower on the CPU, and $4-5$ percentage points better on AP than the original (\cite{cao_affinity_2017}), though it still remains $5\times$ slower than our network on a GPU, and $\approx80\times$ slower than our network on a CPU.

Thus, of the different possible designs of the \textit{SelecSLS} module, and the inter-module skip connectivity choices, the best design for \textit{SelecSLS Net} is the one with concatenation-skip modules, cross-module skip connectivity to the first module in the level, and $n_o$(W) scheme for module outputs. Refer to Section~7.4 in the main manuscript for further comparisons of our architecture against ResNet-50 and ResNet-34 baselines on single-person and multi-person 3D pose benchmarks.

\begin{table}[]
\renewcommand{\tabcolsep}{1.5pt}
\centering
\caption{
Evaluation of possible baseline architecture choices for the core network. The networks are trained on MPI~\shortcite{andriluka_mpii2d_cvpr14} and LSP~\shortcite{johnson_lspet_cvpr11,johnson_lsp_bmvc10} single person 2D pose datasets, and evaluated on LSP testset. 
The inference speed ratios are with respect to ResNet-50 forward pass time for $320\times320$ pixel images on an NVIDIA K80 GPU, using \cite{joli_caffe} with optimized depthwise convolution implementation.
}

\resizebox{0.7\columnwidth}{!}{%
\begin{tabular}{l|c||c}
Core Network                    & ~~~~~PCK~~~~~  & FP Speed Ratio                 \\ \hline
MobileNetV2 1.0x~\shortcite{sandler2018mobilenetv2}               & 85              & 1.78               \\
MobileNetV2 1.3x~\shortcite{sandler2018mobilenetv2}            & 86              & 1.51               \\
Xception~\shortcite{chollet2017xception}            & 81              & 0.67               \\
InceptionV3~\shortcite{szegedy2016rethinking}           & 88              & 0.96               \\
ResNet-34~\shortcite{he_resnet_cvpr2016}             & 89              & 1.27             \\ 
ResNet-50~\shortcite{he_resnet_cvpr2016}            & 89              & 1.00           \\ 
\end{tabular}
}
\label{tbl:lsp_eval}
\end{table}

\begin{table}[t]
\renewcommand{\tabcolsep}{1.5pt}
\centering
\caption{Evaluation of design decisions for first stage of our system. We evaluate different core networks with the 2D pose branch on a subset of validation frames of MS COCO dataset. Also reported are the forward pass timings of the core network and the 2D pose branch on different GPUs (K80, TitanX (Pascal)) as well as Xeon E5-1607 CPU on $512\times320$ pixel input. We also evaluate the publicly available model of \cite{cao_affinity_2017} on the same subset of validation frames. }

\resizebox{\columnwidth}{!}{%
\begin{tabular}{l|ccc|ccc|ccc}
                          & \multicolumn{3}{c|}{\textbf{FP Time }}          & \multicolumn{1}{l}{} & \multicolumn{1}{l}{} & \multicolumn{1}{l|}{} & \multicolumn{1}{l}{} & \multicolumn{1}{l}{} & \multicolumn{1}{l}{} \\
\textbf{Core Network}     & \textbf{K80} & \textbf{TitanX} & \textbf{CPU} & \textbf{AP}          & \textbf{AP\textsubscript{0.5}}      & \textbf{AP\textsubscript{0.75}}      & \textbf{AR}          & \textbf{AR\textsubscript{0.5}}      & \textbf{AR\textsubscript{0.75}}     \\ \hline
ResNet-50                  & 35.7ms       & 9.6ms                    & 349ms                          & 48.8                 & 74.6                 & 52.1                  & 53.2               & 76.8                 & 56.3                \\
ResNet-34                  & 25.7ms       & 5.7ms                    & 269ms                          & 46.4                 & 72.7                 & 47.3                  & 51.3                & 75.2                 & 52.8                 \\ \hline 
\multicolumn{10}{l}{Ours} \\ \hline
~~Add-Skip Prev. (B)             & 24.5ms       & 6.5ms                    & 167ms                          & 47.0                & 73.4                 & 49.7                  & 51.7                 & 75.6                 & 54.5                 \\
~~Conc.-Skip Prev. (B)          & 24.3ms       & 6.3ms                    & 172ms                          & 47.6                 & 73.3                 & 50.7                  & 52.6                 & 76.1                 & 55.6                \\
~~Conc.-Skip Prev. (W) & 25.0ms         & 6.7ms                    & 184ms                          & 48.3                 & 74.4                 & 51.1                    & 52.9                  & 76.5                   & 55.7                 \\
~~Conc.-Skip First (W) & 25.0ms         & 6.7ms                    & 184ms                          & 48.6                 & 74.2                 & 52.2                  & 53.3                 & 76.6                 & 56.7                 \\ \hline \multicolumn{10}{l}{}\\ \hline
\cite{cao_affinity_2017}            & 243ms        & 73.4ms                   & 3660ms                         & 58.0                   & 79.5                 & 62.9                  & 62.1                 & 81.2                 & 66.5                
\end{tabular}
}
\label{tbl:coco_eval}
\end{table}

\section{SelecSLSNet on Image Classification}
\begin{figure}[t]
  \includegraphics[width=0.85\linewidth]{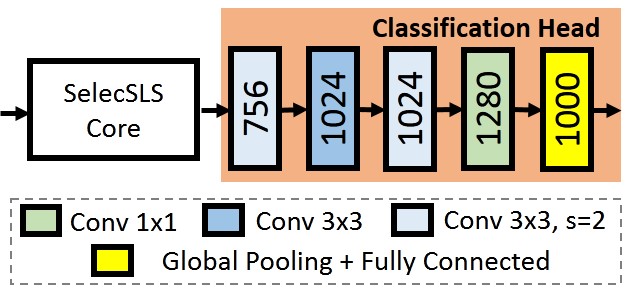}
  \caption
  {\change{For experiments on the image classification task on ImageNet~\cite{imagenet_ijcv2015}, we use the same core architecture design as SelecSLSNet for the multi-person task. Group convolutions are not used in the core network for this task. Inplace of the '2D Branch' and `3D Branch' described in Fig. 2 in the main manuscript, we use a 5 layer network as the classification head. As with the various 2D and 3D pose estimation tasks described previously, the network matches the accuracy of ResNet-50 on ImageNet as well, while being $1.3-1.4\times$ faster.}}

  \label{fig:selecsls_imagenet}
\end{figure}

\change{To demonstrate the efficacy of our proposed architecture on tasks beyond 2D and 3D body pose estimation, we train a variant of SelecSLSNet on ImageNet~\cite{imagenet_ijcv2015}, a large scale image classification dataset. The network architecture is shown in Figure~\ref{fig:selecsls_imagenet}, and the results are shown in Table~\ref{tbl:selecsls_imagenet}.}

\begin{table}[]
\centering
\caption{\change{Results of SelecSLSNet on image classification on the ImageNet dataset. The top-1 and top-5 accuracy on the Imagenet validation set is shown, as well the maximum batch size of $224\times224$ pixel images that can be run in inference mode on an Nvidia V100 16GB card, with FP16 compute. Also shown is the peak throughput obtained with each network, and the batch size of peak throughput (in brackets). SelecSLSNet achieves comparable performance to ResNet-50~\cite{he_resnet_cvpr2016}, while being $1.3-1.4\times$ faster, and with a much smaller memory footprint.}}
\resizebox{0.9\columnwidth}{!}{%
\label{tbl:selecsls_imagenet}
\begin{tabular}{l|c|c|cc|}
          & \textbf{Speed} & \textbf{Maximum}    & \multicolumn{2}{c|}{\textbf{Accuracy}} \\ \cline{4-5} 
          & \textbf{(Images / sec)}    & \textbf{Batch Size} & \textbf{top-1}         & \textbf{top-5}         \\ \hline
ResNet-50 & 2700                     & 1200  (1024)              & 78.5                   & 94.3                   \\
SelecSLSNet  & 3900                     & 2000  (1800)              & 78.4                   & 98.1                  
\end{tabular}
}
\end{table}

\section{Ablation of Input to \textit{Stage II}}
\label{sec:ablation}
We evaluate variants of \textit{Stage II} network taking different subsets of outputs from \textit{Stage I} as input. We compare the \textit{Stage II} output, without \textit{Stage III} on MPI-INF-3DHP 
single person benchmark.
On the single person benchmark (Table~\ref{tbl:mpi_inf_input_ablation}), using only the 2D pose from the 2D branch as input to Stage II, without having trained the 3D branch for Stage I, results in a 3DPCK of 76.0. When using 2D pose from a network with a 3D branch, trained additionally on MuCo-3DHP dataset, we see a minor performance decrease to 75.5 3DPCK. Though it comes with a performance improvement on challenging pose classes such as `Sitting' and `On The Floor' which are under-represrented in MSCOCO. 
Adding other components on top of 2D pose, such as the joint detection confidences $C_k$, and output features from the 3D branch $\{l_{j,k}\}_{j=1}^{J}$ (as described in Section~4.1.2 in the main manuscript) leads to consistent improvement as more components are subsequently used as input to \textit{Stage II}. Using joint detection confidences $C_k$ with 2D pose increases the accuracy to 77.2 3DPCK, and incorporating 3D pose features $\{l_{j,k}\}_{j=1}^{J}$ increases the accuracy to 82.8 3DPCK, and both lead to improvements in AUC and MPJPE as well as improvements for both simpler poses such as upright `Standing/walking' as well as more difficult poses such as `Sitting' and `On the Floor'

This shows the limitations of 2D-3D `lifing' approaches, and demonstrates that incorporating additional information, such as the joint detection confidences, and our proposed 3D pose encoding that uses local kinematic context (channel-sparse supervision) improve the pose performance, leads to significant improvements in 3D pose accuracy. 

\begin{table}[t]
\renewcommand{\tabcolsep}{1.5pt}
\centering
\caption{Evaluation of the impact of the different components from \textit{Stage I} that form the input to \textit{Stage II}. The method is trained for multi-person pose estimation and evaluated on the MPI-INF-3DHP~\shortcite{mehta_mono_3dv17} single person 3D pose benchmark. The components evaluated are the 2D pose predictions $P_{k}^{2D}$, the body joint confidences $C_k$, and the set of extracted 3D pose encodings $\{l_{j,k}\}_{j=1}^{J}$. 
Metrics used are: 3D percentage of correct keypoints (\textbf{3DPCK}, higher is better), area under the curve (\textbf{AUC}, higher is better) and mean 3D joint position error (\textbf{MJPE}, lower is better). Also shown are the results with channel-dense supervision of 3D pose encodings $\{l_{j,k}\}_{j=1}^{J}$, as well as evaluation of \textit{Stage III} output.}

\resizebox{0.9\columnwidth}{!}{%
\begin{tabular}{ll|c|c|c|cc}

                     &  & \multicolumn{3}{c|}{\textbf{3DPCK}}             & \multicolumn{1}{l}{}                & \multicolumn{1}{l}{}  \\ \cline{3-5}
\textbf{Stage II}    &  & \textbf{Stand} & \textbf{}    & \textbf{On The} & \multicolumn{2}{c}{\textbf{Total}}                                                \\\cline{6-7}
\textbf{Input}       &  & \textbf{/Walk} & \textbf{Sitt.} & \textbf{Floor}  & \multicolumn{1}{c|}{\textbf{3DPCK}} & \textbf{AUC}                \\ \hline

$P_{k}^{2D}$  (2D Branch Only)     \T \B    &  & 86.4           & 76.3         & 44.9            & \multicolumn{1}{c|}{76.0}           & 42.1                                 \\ \hline
$P_{k}^{2D}$ \T                 &  & 79.8           & 78.4         & 58.5            & \multicolumn{1}{c|}{75.5}           & 41.3                                 \\
$P_{k}^{2D} + C_k$ \T          &  & 85.9           & 79.4         & 58.7            & \multicolumn{1}{c|}{77.2}           & 42.2                                 \\
$P_{k}^{2D} + C_k + \{l_{j,k}\}_{j=1}^{J} $\T   \B     &  & \textbf{88.4}           & \textbf{85.8}         & \textbf{70.7}            & \multicolumn{1}{c|}{\textbf{82.8}}           & \textbf{45.3}                                  \\ \hline
\multicolumn{7}{l}{Channel-Dense $\{l_{j,k}\}_{j=1}^{J}$ Supervision \B} \\  \hline
$P_{k}^{2D} + C_k + \{l_{j,k}\}_{j=1}^{J} $ \T        &  & {87.0}           & 83.6         & 61.5            & \multicolumn{1}{c|}{80.1}           & 43.3                                 \\ \hline 
\end{tabular}
}
\label{tbl:mpi_inf_input_ablation}
\end{table}

\section{Sequential Motion Capture (\textit{Stage III}): Additional Details}
\paragraph{Absolute Height Calibration}

As mentioned in the main document, to allow more accurate camera relative localization, we can optionally utilize the ground plane as reference geometry. 
First, we determine the camera relative position of a person by shooting a ray from the camera origin through the person's foot detection in 2D and computing its intersection with the ground plane. The subject height, $h_k$, is then the distance from the ground plane to the intersection point of a virtual billboard placed at the determined foot position and the view ray through the detected head position.
Because we want to capture dynamic motions such as jumping, running, and partial (self-)occlusions, we cannot assume that the ankle is visible and touches the ground at every frame. Instead, we use this strategy only once when the person appears. As shown in the accompanying video, such a height calibration strategy allows reliable camera-relative localization of subjects in the scene even when they are not in contact with the ground plane.

In practice, we compute intrinsic and extrinsic camera parameters once prior to recording using checkerboard calibration. Other object-free calibration approaches would be feasible alternatives~\cite{yang2018recovering,zanfir2018monocular}.

\paragraph{Inverse Kinematics Tracking Error Recovery:}
Since we use gradient descent for optimizing the fitting energy, we can monitor the gradients of $E_{3D}$ and $E_{lim}$ terms in $\mathcal{E}(\theta_1[t], \cdots, \theta_K[t])$ to identify when tracking has failed, either due to a failure to correctly match subjects to tracks because of similar appearance and pose, or when the fitting gets stuck in a local minimum. When the gradients associated with these terms exceed a certain threshold for a subject for 30 frames, the identity and pose track of the subject is re-initialized.

\begin{figure}[h!]
  \includegraphics[width=0.95\linewidth]{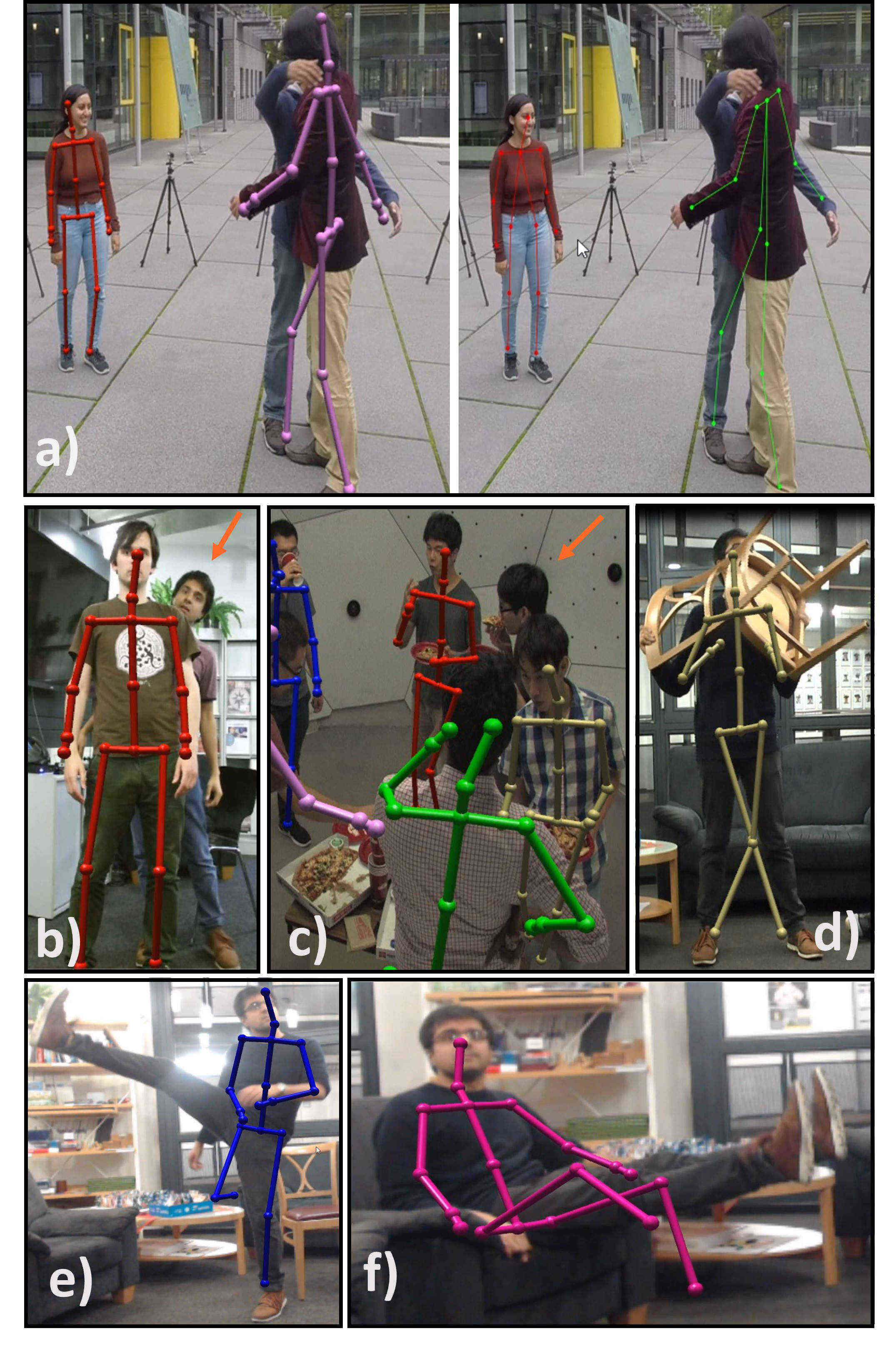}
  \caption{\textbf{Limitations:} a)3D pose inaccuracy due to 2D pose limb confusion, b),c) Person not detected due to neck occlusion and extreme occlusion, d) Body orientation confusion due to occluded face e),f) Unreliable pose estimates for poses drastically different from the training data.}
  \label{fig:failure_cases}
\end{figure}

\begin{figure*}[t]
  \includegraphics[width=\linewidth]{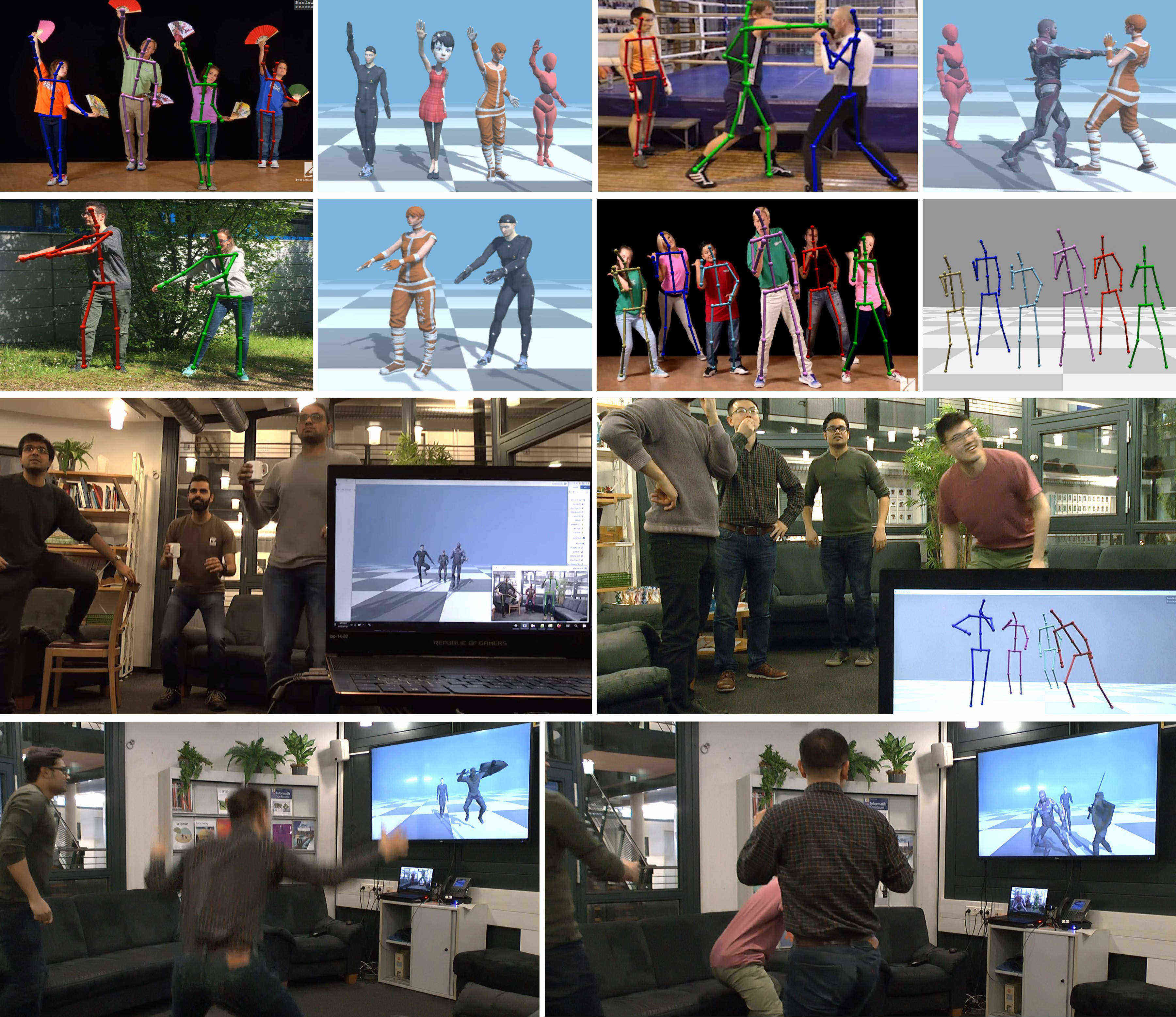}
  \caption
  {\textbf{Live Interaction and Virtual Character Control:} The temporally smooth joint angle predictions from \textit{Stage III} can be readily employed for driving virtual characters in real-time.
  The top shows our system driving virtual skeletons and characters in real-time with the captured motion. On the bottom, our system is set up as a Kinect-like game controller, allowing subjects to interact with their virtual avatars live. 
  Some images courtesy Boxing School Alexei Frolov (\url{https://youtu.be/dbuz9Q05bsM}), and Music Express Magazine (\url{https://youtu.be/kX6xMYlEwLA}, \url{https://youtu.be/lv-h4WNnw0g}).
  Please refer to the supplemental video for more results.  
  }
  \label{fig:character_control_supp}
\end{figure*}

\begin{figure*}[t]
  \includegraphics[width=1.0\linewidth]{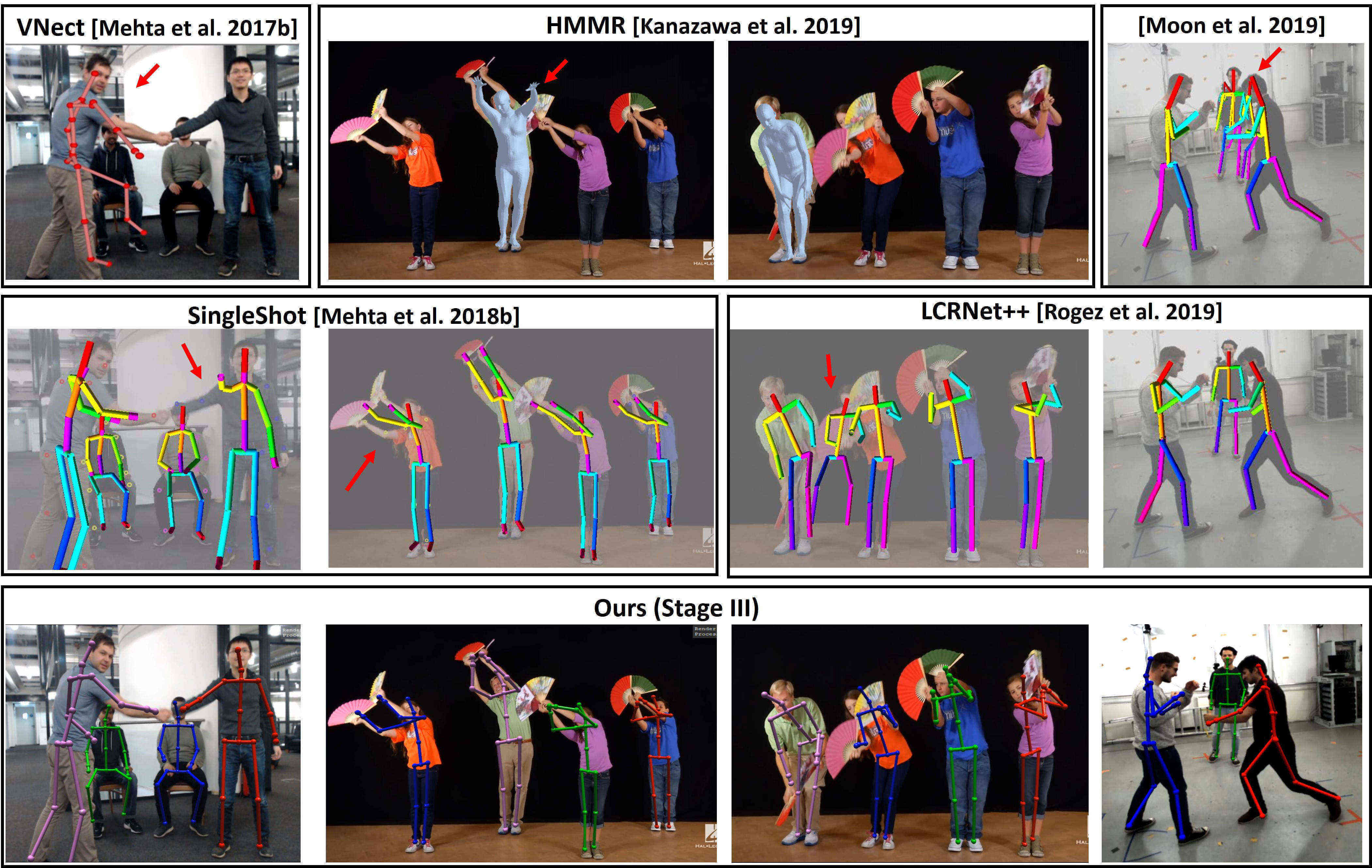}
  \caption
  {Real-time 3D Pose approaches such as VNect~\shortcite{VNect_SIGGRAPH2017} only work in single person scenarios, and not designed to be occlusion robust or to handle other subjects in close proximity to the tracked subject. LCRNet++~\shortcite{rogez_lcrpp} is able to handle multi-person scenarios, but works at interactive frame rates, and requires post processing to be able to fuse the multiple pose proposals generated per subject. The post-processing is not always successful at fusing all proposals, leading to ghost predictions. The offline single-person approach HMMR~\shortcite{humanMotionKanazawa19} uses 2D multi-person pose estimation as a pre-processing step and is thus able to handle unoccluded subjects in multi-person scenes in a top-down way. However, the approach fails under occlusion, and the run-time scales linearly with the number of subjects in the scene. The multi-person approach of \cite{mehta_3dv18} jointly handles multiple subjects in the scene, however shows failures in cases of inter-personal proximity. The multi-person approach of \cite{Moon_2019_ICCV_3DMPPE} works offline, and similar to LCRNet++ it often produces spurious predictions due to the difficulty of filtering multiple proposals from top-down approaches. Here for our bottom-up approach (bottom), we show the 3D skeleton from \textit{Stage III} reprojected on the image.
  Some images courtesy Music Express Magazine (\url{https://youtu.be/kX6xMYlEwLA}).
  }
  \label{fig:vnect_comparisons}
\end{figure*}

\section{More Qualitative Results}
\parahead{Limitations}
Figure~\ref{fig:failure_cases} shows visual examples of the limitations of our approach, as discussed in Section 8 in the main manuscript. Our system mispredicts when the underlying 2D pose prediction mispredicts limb associations across subject. When the neck of the subject is occluded, we treat the subject as not present, even when the rest of the body is visible. This could be handled by using multiple reference joints on the body, instead of just the neck. 
Also, as our approach is a learning based approach, it mispredicts when the presented pose is outside the training distribution. 

\parahead{Comparisons With Prior Work}
Figure~\ref{fig:vnect_comparisons} shows visual comparisons of our approach to prior single-person and multi-person approaches. Also refer to the accompanying video for further comparisons. Results of our real-time system are comparable or better in quality than both, single-person and multi-person approaches, many of which run at offline~\cite{Moon_2019_ICCV_3DMPPE,humanMotionKanazawa19,kanazawa2018endtoend,mehta_3dv18} and interactive~\cite{rogez_lcrpp,dabral2019multi} frame-rates. As shown in the video, the temporal stability of our approach is comparable to real-time~\citep{VNect_SIGGRAPH2017} and offline~\cite{humanMotionKanazawa19} temporally consistent single-person approaches. Our approach differs from much of recent multi-person approaches in that ours is a bottom-up approach, while others employ a top-down formulation~\cite{rogez_lcrpp,rogez_lcr_cvpr17,dabral2019multi,Moon_2019_ICCV_3DMPPE} inspired by work on object detection. Top-down approaches produce multiple predictions (proposals) per subject in the scene, and require a post-processing step to filter. Even when carefully tuned, this filtering step can either suppress valid predictions (two subjects with similar poses in close proximity) or fail to suppress invalid predictions (ghost predictions where there is no subject in the scene).

\parahead{Live Interaction and Character Control}
Figure~\ref{fig:character_control_supp} shows additional examples of live character control with our real-time monocular motion capture approach. Also refer to the accompanying video for more character control examples. Our system can act as a drop-in replacement for typical depth sensing based game controllers, allowing subjects to interact with their live avatars. 

\begin{figure*}[t]
  \includegraphics[width=1.0\linewidth]{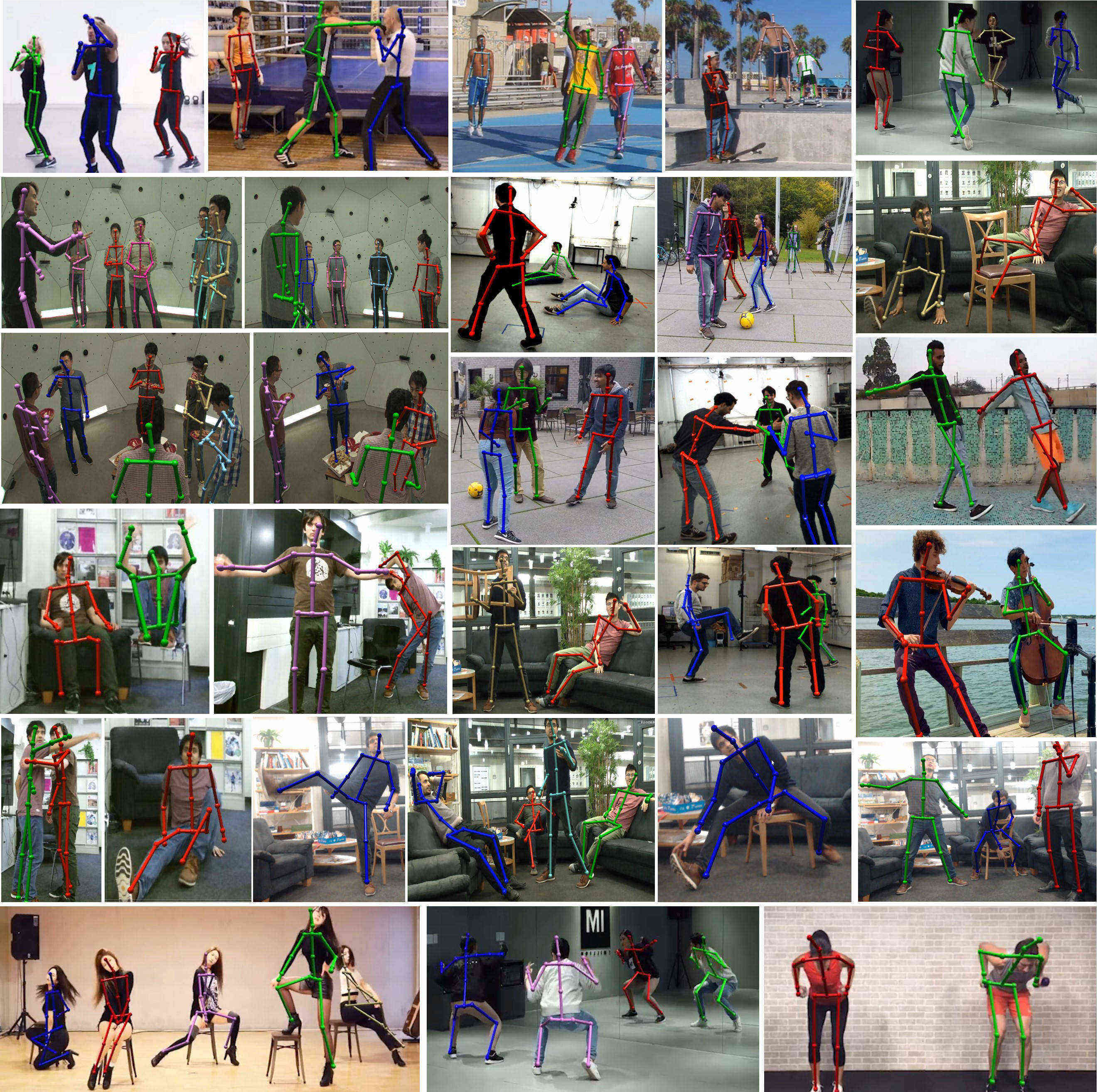}
  \caption
  {Monocular 3D motion capture results from our full system (\textit{Stage III}) on a wide variety of multi-person scenes, including Panoptic~\shortcite{joo_panoptic_iccv2015} and MuPoTS-3D~\shortcite{mehta_3dv18} datasets. Our approach handles challenging motions and poses, including interactions and cases of self-occlusion. 
  Some images courtesy KNG Music (\url{https://youtu.be/_xCKmEhKQl4}), 1MILLION TV (\url{https://youtu.be/9HkVnFpmXAw}), Indian dance world (\url{https://youtu.be/PN6tRmj6xGU}), 7 Minute Mornings (\url{https://youtu.be/oVgG5ENXyVs}), Crush Fitness (\url{https://youtu.be/8qFwPKfllGI}), Boxing School Alexei Frolov (\url{https://youtu.be/dbuz9Q05bsM}), and Brave Entertainment (\url{https://youtu.be/ZhuDSdmby8k}).
  Please refer to the accompanying video for more results.}
  \label{fig:more_results_supp}
\end{figure*}

\parahead{Diverse Pose and Scene Settings}
Figure~\ref{fig:more_results_supp} shows the 3D capture results from our system (\textit{Stage III}) overlaid on input images from diverse and challenging scenarios. See the accompanying video for additional results. Our approach can handle a wide range of poses, in a wide variety of scenes with different lighting conditions, background, and person density. 

\clearpage
\bibliographystyle{ACM-Reference-Format}
\bibliography{article_eccv,article}

\end{document}